\documentclass[fleqn,10pt]{wlscirep}
\usepackage[utf8]{inputenc}
\usepackage[T1]{fontenc}
\usepackage{amsmath,amssymb}
\usepackage{dcolumn,setspace}
\usepackage{graphics}
\usepackage{rotating}
\usepackage[graphicx]{realboxes}
\usepackage{float}
\usepackage{booktabs}
\usepackage{makecell}
\usepackage{listings}

\usepackage{url}
\usepackage{enumitem}
\usepackage{float}
\usepackage{multicol}
\usepackage{amsmath}
\usepackage[linesnumbered,ruled]{algorithm2e}
\usepackage[scaled=0.85]{beramono}

\makeatletter
\newcommand\xleftrightarrow[2][]{%
  \ext@arrow 9999{\longleftrightarrowfill@}{#1}{#2}}
\newcommand\longleftrightarrowfill@{%
  \arrowfill@\leftarrow\relbar\rightarrow}
\makeatother

\DeclareMathAlphabet{\pazocal}{OMS}{zplm}{m}{n}
\newcommand{\symA}{\pazocal{A}}
\newcommand{\symE}{\pazocal{E}}
\newcommand{\symR}{\pazocal{R}}

\newcommand{\symNG}{\pazocal{NG}}

\newcommand{\symLD}{\pazocal{LD}}

\newcommand{\symS}{\pazocal{S}}
\newcommand{\symC}{\pazocal{C}}

\newcommand{\symG}{\pazocal{G}}
\newcommand{\symL}{\pazocal{L}}

\newcommand{\symSE}{\pazocal{SE}}
\newcommand{\symP}{\pazocal{P}}
\newcommand{\symT}{\pazocal{T}}
\newcommand{\symM}{\pazocal{M}}
\DeclareMathAlphabet{\pazocal}{OMS}{zplm}{m}{n}

\title{An Empirical Meta-analysis of the Life Sciences (Linked?) Open Data on the Web}

\author[1,*,+]{Maulik R. Kamdar}
\author[1]{Mark A. Musen}
\affil[1]{Center for Biomedical Informatics Research, Stanford University, Stanford, CA, USA}

\affil[*]{maulik@maulik-kamdar.com}

\affil[+]{The author is currently affiliated with Elsevier Health Markets, Philadelphia, PA, USA.}

\begin{abstract}
While the biomedical community has published several ``open data'' sources in the last decade, most researchers still endure severe logistical and technical challenges to discover, query, and integrate heterogeneous data and knowledge from multiple sources. To tackle these challenges, the community has experimented with Semantic Web and linked data technologies to create the Life Sciences Linked Open Data (LSLOD) cloud. In this paper, we extract schemas from more than 80 publicly available biomedical linked data graphs into an LSLOD schema graph and conduct an empirical meta-analysis to evaluate the extent of semantic heterogeneity across the LSLOD cloud. We observe that several LSLOD sources exist as stand-alone data sources that are not inter-linked with other sources, use unpublished schemas with minimal reuse or mappings, and have elements that are not useful for data integration from a biomedical perspective. We envision that the LSLOD schema graph and the findings from this research will aid researchers who wish to query and integrate data and knowledge from multiple biomedical sources simultaneously on the Web.
\end{abstract}

\begin{document}

\flushbottom
\maketitle

\thispagestyle{empty}

\section*{Introduction}

Over the last decade, the biomedical research community has published and made available, on the Web, several sources consisting of biomedical data and knowledge: anonymized medical records \cite{johnson2016mimic}, imaging data \cite{clark2013cancer}, sequencing data \cite{weinstein2013cancer}, biomedical publications \cite{pubmedlink}, biological assays \cite{fu2015pubchemrdf}, chemical compounds and their activities \cite{hastings2013chebi}, biological molecules and their characteristics \cite{uniprot2008universal}, knowledge encoded in biological pathways \cite{croft2014reactome,kanehisa2000kegg}, animal models \cite{mungall2017monarch}, drugs and their protein targets \cite{wishart2006drugbank}, medical knowledge on organs, symptoms, diseases, and adverse reactions \cite{bodenreider2008biomedical}, etc. However, biomedical researchers still face severe logistical and technical challenges to integrate, analyze, and visualize heterogeneous data and knowledge from these diverse and often isolated sources. Researchers need to be aware of the availability and accessibility of these sources on the Web, and need extensive computational skills and knowledge to query and explore them. To systematically consume and integrate the data and knowledge from these sources for use in their analyses, researchers often spend a significant amount of their time dealing with the heterogeneous syntaxes, structures, formats, schemas, and biomedical entity notations used across these sources. 

The complexity and time in inter-disciplinary scientific research drastically increase as the biomedical researcher ends up learning multiple systems, configurations, and means to reconcile similar entities across different sources. In most cases, the biomedical researcher just wishes to retrieve relevant data pertaining to their unique criteria or to retrieve answers to specific queries, such as \textit{``What are the medications prescribed to melanoma patients who have a V600E mutation in their BRAF gene?''}, or \textit{``List molecular characteristics of antineoplastic drugs that target Estrogen Receptor $\alpha$ and have molecular weight less than 300 g/mol''}. However, due to the current state of the biomedical data and knowledge landscape with multiple, isolated, heterogeneous sources, the researcher has to hop across several Web portals (e.g., Pubchem \cite{fu2015pubchemrdf}) and search engines (e.g., PubMed \cite{pubmedlink}) for these tasks.

Semantic Web and linked data technologies are deemed to be promising solutions to tackle these challenges, and enable toward Web-scale data and knowledge integration and semantic processing in several biomedical domains, such as pharmacology and cancer research \cite{kamdar2019enabling}. Biomedical researchers have been some of the earliest adopters of these technologies, and have used them to drive knowledge management, semantic search, clinical decision support, rule-based decision making, enrichment analysis, data annotation, and data integration. Moreover, these technologies are used to represent data and knowledge stored in isolated, heterogeneous sources on the Web to create a network of ``linked knowledge graphs'', often referred to as the Life Sciences Linked Open Data (LSLOD) cloud \cite{kamdar2019enabling}. A diagrammatic representation of the LSLOD cloud can be found as part of the popular Linked Open Data cloud diagram\footnote{\url{https://lod-cloud.net/}}. These technologies have matured enough that they are also being widely adopted by several companies, and subsets of the generated knowledge graphs are used in biomedical applications (e.g., Google \cite{ramaswami2015remedy}, IBM \cite{aocnp2015watson}, Elsevier \cite{kamdar2020text},  NuMedii \cite{dastgheib2018accelerating}, Pfizer \cite{pfizerlink}, etc.). Semantic Web technologies could also be combined with natural language processing-based relation extraction methods to systematically extract scientific relations between biomedical entities and to make them available for querying and consumption through a structured representation of the knowledge graph \cite{percha2018global,kamdar2020text}.

Ideally, the LSLOD cloud should serve as an open Web-based scalable infrastructure through which biomedical researchers can query, retrieve, integrate, and analyze data and knowledge from multiple sources, without the requirement on the part of the researchers to download and manually integrate those sources and be concerned with the location, heterogeneous schemas, syntaxes, varying entity notations, representations, or the mappings to reconcile similar concepts, relations, and entities between these sources. However, it is well known that the LSLOD cloud has several shortcomings for this vision to be achieved. Specifically, most biomedical researchers find it difficult to query and integrate data and knowledge from the LSLOD cloud for use in their downstream applications, since the LSLOD cloud faces a set of challenges similar to those that it had set out to solve in the first place \cite{kamdar2019enabling}. The rampant semantic heterogeneity across the sources in the LSLOD cloud, due to the heterogeneous schemas, the varying entity notations, the lack of mappings between similar entities, and the lack of reuse of common vocabularies, leads to the question on whether the LSLOD cloud can truly be considered \textit{``linked''}.

In this paper, we present our results on an empirical meta-analysis conducted on more than 80 data and knowledge sources in the Life Sciences Linked Open Data (LSLOD) cloud. The main contributions of this research can be outlined as follows:
\begin{enumerate}[noitemsep]
    \item Using an approach to automatically extract schemas and vocabularies from a set of publicly available linked data endpoints that expose biomedical data and knowledge sources on the Web, we have created an LSLOD schema graph encapsulating content from more than 80 sources.
    \item We estimate the extent of reuse of content in LSLOD sources from popular ontologies and vocabularies, as well as the level of intra-linking and inter-linking across different LSLOD sources.
    \item Combining methods of similarity computation and community detection based on word embeddings, we identify communities of similar content across the LSLOD cloud.
\end{enumerate}

We believe that the resources and findings established through this meta-analysis will aid both: \emph{i)} the biomedical researchers, who wish to query and use data and knowledge from LSLOD cloud in their applications, and \emph{ii)} the Semantic Web researchers, who can develop methods and tools to increase reuse and reduce semantic heterogeneity across the LSLOD cloud, as well as develop efficient federated querying mechanisms over the LSLOD cloud to handle heterogeneous schemas and diverse entity notations. In subsequent sections of this paper, we provide a brief overview on the Semantic Web technologies and the Life Sciences Linked Open Data (LSLOD) cloud, delve deeper into the different biomedical data and knowledge sources that were used in this research, and provide an overview on the different methods used to extract schemas and vocabularies from publicly available linked data graphs and to evaluate the reuse and similarity of content across the LSLOD cloud. We will present the results from this empirical meta-analysis of the quality, reuse, linking, and heterogeneity across the LSLOD cloud, summarize some of our key findings and observations on the state of biomedical linked data on the Web, and discuss what it means moving forward on the querying and consumption of data and knowledge on the Web in biomedical applications, in a more systematic and integrated fashion. The LSLOD schema graph and other pertinent data, as well as the results and visualizations from this meta-analysis are made available online at \url{http://onto-apps.stanford.edu/lslodminer}.

\section*{Background and Related Work}
\label{sect:background}
\subsection*{Semantic Web Technologies and Life Sciences Linked Open Data}
\label{intro:semweb}

The Linked Open Data cloud was conceived from a vision that a decentralized, distributed, and heterogeneous data space, extending over the traditional Web (i.e., World Wide Web, or Web of Documents), can reveal hidden associations that were not directly observable  \cite{berners2001semantic}. To achieve this vision, the Semantic Web community has developed several standards, languages, and technologies that aim to provide a common framework for data and knowledge representation --- linking, sharing, and querying, across application, enterprise, and community boundaries. Semantic Web languages and technologies have been used to represent and link data and knowledge sources from several different fields such as life sciences, geography, economics, media, and statistics, to essentially create a linked network of these sources and to provide a scalable infrastructure for structured querying of multiple heterogeneous sources simultaneously, for Web-scale computation, and for seamless integration of big data. 

Some Semantic Web languages and technologies have been standardized and recommended by the World Wide Web Consortium (W3C) and are also widely used by the biomedical community to create the Life Sciences Linked Open Data (LSLOD) cloud. For example, The Resource Description Framework (RDF), a simple, standard triple-based model for data interchange on the Web, is used to represent information resources (e.g., biomedical entities, relations, classes) as linked graphs on the LSLOD cloud \cite{klyne2006resource}. Each resource is uniquely represented using a HTTP Uniform Resource Identifier (URI), so that users can view information pertaining to that resource using a Web browser (i.e., Web-dereferenceable). An example of an HTTP URI from the linked data graph of DrugBank --- a knowledge base of drugs and their protein targets  \cite{wishart2006drugbank} ---  is \url{http://bio2rdf.org/drugbank:DB00619}, where \url{http://bio2rdf.org/drugbank:} is considered to be the URI namespace -- a declarative region that provides a scope to the identifiers, and \texttt{DB00619} is the specific identifier of the drug \textsc{Gleevec}. RDF extends the linking structure of the Web by using the URIs to represent relations between two resources. Hence, RDF facilitates integration and discovery of relevant data and knowledge on the Web even if the schemas and syntaxes of the underlying data sources differ. 

The vocabularies and schemas of biomedical RDF graphs are generally annotated by the elements from the Resource Description Framework Schema (RDFS) language \cite{mcbride2004resource}. Similarly, in the last few years, the biomedical community has adopted the Web Ontology Language (OWL) as the consensus knowledge representation language to develop biomedical ontologies \cite{bechhofer2009owl}. RDFS vocabularies and OWL ontologies enable the modeling of any domain in the world by explicitly describing existing entities, their attributes and the relationships among these entities. These entities and relationships are usually classified in specialization or generalization hierarchies. An ontology also contains logical definitions of its terms, along with additional relations, to facilitate advanced search functionalities \cite{gruber1995toward}. Whereas, RDF is essentially only a triple-based, schema-less modeling language, semantics expressed in RDFS vocabularies and OWL ontologies can be exploited by computer programs, called reasoners, to verify the consistency of assertions in an RDF graph and to generate novel inferences \cite{kamdar2019enabling}.

After the data and knowledge sources have been published as linked data using the RDF triple-based model, with semantics encoded using the RDFS and OWL languages, the SPARQL graph query language can be used to query across multiple diverse sources using federated architectures \cite{prud2008sparql,kamdar2017phlegra,saleem2014big}. While this distributed querying approach is inspired from the relational database community, SPARQL query federation architectures leverage the advantages provided by the graphical, uniform syntax, and schema-less nature of RDF to achieve query federation with minimal effort.  

\subsubsection*{Representative Example on Use of Semantic Web Technologies in Biomedicine}

\begin{figure}[!htb]
\begin{centering}
\includegraphics[scale=0.23]{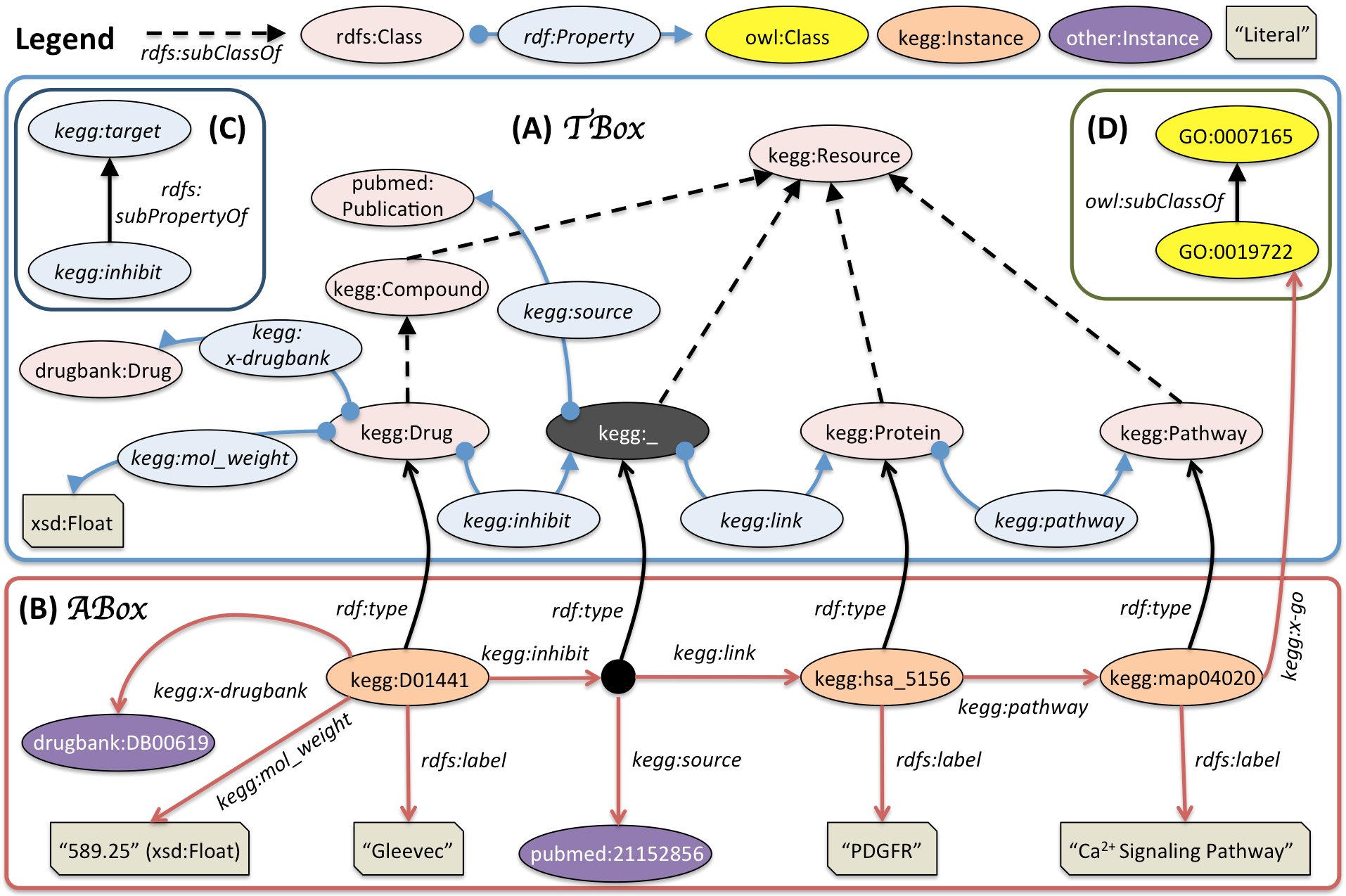}
\par\end{centering}
\caption[Diagrammatic representation to depict annotated RDF graphs]{\textbf{Diagrammatic representation to depict annotated RDF graphs.} The KEGG RDF graph whose vocabulary is annotated using RDFS constructs and classes from the Gene Ontology. \textbf{A)} The $\symT$-Box consists of the terminological component of the KEGG RDF Graph (i.e., the vocabulary). The pink nodes represent the different classes from KEGG and other RDF graph vocabularies (e.g., \texttt{drugbank:Drug}) annotated with \texttt{rdfs:Class} construct. The blue nodes represent the different properties annotated with \texttt{rdf:Property}  (with associated domain and range classes). \textbf{B)} The $\symA$-Box consists of the assertional component of the KEGG RDF graphs, with different individuals (e.g., \texttt{kegg:D01441}) explicitly typed with corresponding classes from $\symT$-Box. \textbf{C)} The $\symR$-Box (relational component) consists of the property hierarchy. \textbf{D)} The KEGG individual (pathway  \textsc{Ca}$^{2+}$ \textsc{Signaling Pathway})) may be cross-linked to the corresponding Gene Ontology class, causing a semantic mismatch (i.e., a class is considered equivalent to an individual).}
\label{fig:rdfsowl}
\end{figure}

Using \textbf{Figure \ref{fig:rdfsowl}}, we give a more concrete real-world example on the use of Semantic Web technologies for representing data in the Kyoto Encyclopedia of Genes and Genomes (KEGG) --- an integrated data source consisting of several databases, broadly categorized into biological pathways, proteins, and drugs \cite{kanehisa2000kegg} -- and linking the different biomedical entities to similar entities in other biomedical sources, such as DrugBank \cite{wishart2006drugbank}, PubMed \cite{pubmedlink}, and Gene Ontology --- a unified representation of gene and gene product attributes across all species \cite{ashburner2000gene}.

\texttt{kegg:Drug}\footnote{\texttt{kegg:Drug} is a concise representation for a URI \texttt{http://bio2rdf.org/kegg:Drug}.}, \texttt{kegg:Protein}, and \texttt{kegg:Compound} are all RDFS classes in the KEGG vocabulary (\textbf{Figure \ref{fig:rdfsowl}A}). \texttt{kegg:Drug} is a subclass of \texttt{kegg:Compound} (shown using \texttt{rdfs:subClassOf}). \texttt{kegg:inhibit} and \texttt{kegg:target} are both RDFS properties, with \texttt{kegg:inhibit} being a sub-property of \texttt{kegg:target} (shown using \texttt{rdfs:subPropertyOf} in \textbf{Figure \ref{fig:rdfsowl}C}). Using a reasoning-enabled query engine, a user who queries for \texttt{kegg:Compound} entities should retrieve all \texttt{kegg:Drug} entities, as well as all non-drug compounds. Similarly, all queries that look for \texttt{kegg:target} relations should retrieve all \texttt{kegg:inhibit} relations, along with non-inhibiting interactions. The domain and range of the properties are also shown using the $\bullet$ and $\blacktriangleright$ arrow heads respectively. The KEGG RDF Graph can be aligned to this KEGG RDFS vocabulary, with a class (e.g., \texttt{kegg:Drug}) instantiated with corresponding instances (e.g., \texttt{kegg:D01441} (\textsc{Gleevec})) using the \texttt{rdf:type} construct, and properties realized as property assertions between different instances (e.g., \texttt{kegg:pathway} assertion between \texttt{kegg:hsa\_5156} (\textsc{PDGFR}) and \texttt{kegg:map04020} (\textsc{Ca}$^{2+}$ \textsc{Signaling Pathway})). These property assertions are shown using red arrows (\textbf{Figure \ref{fig:rdfsowl}B}). 

\textbf{Figure \ref{fig:rdfsowl}B} also shows an example of RDF \textit{reification}, where a \textit{blank node} (represented using a black colored node) is used to capture and represent additional information on the interaction between the drug \textsc{Gleevec} and the receptor protein \textsc{PDGFR}, involved in regulating cell growth and implicated in cancer. The \textit{blank node} links the additional information to a publication with a URI from a PubMed linked data graph. Object properties (e.g., \texttt{kegg:pathway}) generally have \texttt{rdfs:Class} classes as domains and ranges, whereas data properties (e.g., \texttt{kegg:mol\_weight}) link an \texttt{rdfs:Class} to a \textit{datatype} (e.g., \texttt{xsd:Float} is the datatype for float numbers).

RDF graphs, annotated using the RDFS language, can be divided into two main components: the terminological component $\symT$-Box and the assertional component $\symA$-Box (\textbf{Figure \ref{fig:rdfsowl}}). The $\symT$-Box consists of only the axioms pertaining to the vocabulary or schema (e.g. \texttt{rdfs:Class}, \texttt{rdfs:subClassOf} axioms, along with other logical combination constructs). The $\symA$-Box consists of axioms pertaining to the individuals and their property assertions. RDF graphs (and their associated vocabularies) can be exposed through SPARQL endpoints on the LSLOD cloud for programmatic access.

In general, classes, properties, and instances in RDF graphs may be cross-linked with elements from external ontologies and vocabularies, and making such linkages is often considered a best practice when modeling and publishing datasets as Linked Open Data \cite{marshall2012emerging}. For example, the pathway \textsc{Ca2+ Signaling Pathway} is represented as a class in the Gene ontology (\texttt{GO:0019722}), but it is represented as an individual (instance) of the \texttt{kegg:Pathway} class in the KEGG RDF Graph (\texttt{kegg:map04020}) in \textbf{Figure \ref{fig:rdfsowl}C,D}. However, due to the simplistic triple-based nature of RDF, the KEGG individual can be cross-linked to the Gene Ontology class (this particular situation is often called semantic mismatch, where a class is considered equivalent to an individual).

\subsection*{Semantic Heterogeneity across Linked Data Graphs}
\label{sect:revocsuc}
Entity reconciliation and integrated querying are major problems in biomedicine, as there is often no agreement on a unique representation of a given entity or a class of entities. Many biomedical entities are referred to by multiple labels, and the same labels may be used to refer to different entities. For example, different gene entities could be represented using Ensembl \cite{yates2016ensembl}, Entrez Gene \cite{maglott2005entrez}, and Hugo Gene Nomenclature Committee (HGNC) \cite{gray2015genenames} identifiers.

One of the key principles for publishing linked data (LD) is the use of HTTP Uniform Resource Identifiers (URIs) for entity reconciliation and integrated querying. Each entity (e.g., \textsc{Lepirudin}), or a class (e.g., \textsc{Drug}), should be represented uniquely using one unique URI and other publishers should reuse that URI in their datasets. However, most data publishers create their own URIs by combining the entity IDs from the underlying data sources (e.g., \textsc{Gleevec} DrugBank ID DB00619) with a unique namespace (e.g. \url{http://bio2rdf.org/drugbank:}) to represent entities. To resolve the problem of entity reconciliation, efforts such as Bio2RDF \cite{callahan2013bio2rdf} and Linking Open Drug Data \cite{jentzsch2009linking} have released guidelines for using cross-reference \texttt{x-ref} attributes rather than using the same URI. Similar entities in different sources should be mapped to each other (e.g., \url{drugbank:DB00619} $\xleftrightarrow[x-ref]{}$ \url{kegg:D01441} to map \textsc{Gleevec} entities in DrugBank and KEGG), or all similar entities should be mapped to a common terminology (e.g., \url{drugbank:BE0000048} $\xleftrightarrow[x-ref]{}$ \url{hgnc:3535} $\xleftrightarrow[x-ref]{}$ \url{kegg:HSA_2147}, where the different URIs for protein \textsc{Prothrombin} are mapped to the Hugo Gene Nomenclature Committee (HGNC) identifier) using such \texttt{x-ref} attributes from popular RDFS vocabularies \cite{marshall2012emerging}.

Ideally, in the case of RDF datasets, the terminological $\symT$-Box elements (i.e., classes and properties used in the schemas) should be reused from existing OWL ontologies or RDFS vocabularies. Large repositories of existing RDFS vocabularies are developed, either curated manually (e.g., Linked Open Vocabularies \cite{lov}) or by exploring the LSLOD cloud through automated crawlers. These repositories can serve as sources for data publishers to reuse elements from existing vocabularies. Moreover, in biomedicine, the classes from existing biomedical ontologies in the BioPortal repository \cite{whetzel2011bioportal} can also be reused when publishing LD. RDF instances in the assertional $\symA$-Box may also be linked to classes from OWL ontologies using equivalence properties, such as \texttt{x-ref} or \texttt{owl:sameAs} as shown in \textbf{Figure \ref{fig:rdfsowl}D}. However, cases where RDF `instances' are mapped to `classes' in other ontologies or LD graphs using equivalence properties are often designated as \textit{semantic mismatch} cases.

In some cases, the same relation may be expressed in different RDF graphs using different semantics or graph patterns \cite{kamdar2019enabling}. For example, the same attribute of molecular weight is expressed in two different LD sources, DrugBank and KEGG, using \texttt{drugbank:molecular-weight} and \texttt{kegg:mol\_weight} respectively. Ideally, the molecular weight attribute assertions should be structured using a uniform property across different sources. This property can originate from any of the LOV vocabularies or biomedical ontologies (e.g. \texttt{CHEMINF:000216} ``average molecular weight descriptor'' from the Chemical Information Ontology). However, in the current state of the LSLOD cloud, a user who wishes to retrieve and reconcile such relations from multiple linked data graphs must be aware of the underlying semantics and the data model used in these graphs. 

There are several benefits for reuse or cross-linking between similar entities or classes in different linked data graphs or ontologies \cite{kamdar2017systematic,marshall2012emerging} --- \emph{i)} reduction in data and knowledge engineering costs since the publisher can reuse existing minted URIs or link to existing entity descriptions, \emph{ii)} enabling semantic interoperability among different datasets and applications, and \emph{iii)} querying multiple linked data graphs simultaneously using federated architectures. The lack of reuse or cross-linking as well as the use of different semantics or graph patterns for modeling data in a given linked data graph leads to an increase of \textit{semantic heterogeneity} across the LSLOD cloud.

\subsection*{Related Research}

Previously, we had conducted a similar study to estimate \textit{term reuse} (i.e., reuse of classes from existing ontologies) and \textit{term overlap} (i.e., similar classes across ontologies which are not reused) across biomedical ontologies stored in the BioPortal repository \cite{kamdar2017systematic}. We found that ontology developers seldom reuse content from existing biomedical ontologies. However, there is significant overlap across the BioPortal ontologies. The concepts of \textit{term reuse} and \textit{term overlap} also hold true in the context of linked data (LD) sources. However, in the biomedical domain, there are several critical differences between ontologies and LD sources due to the manner through which these sources are created and formalized. Some of these differences include: \emph{i)} the vocabularies and schema for biomedical LD sources are significantly smaller compared to most biomedical ontologies, \emph{ii)} LD sources have a much higher number of instances compared to biomedical ontologies and these instances may be reused from external resources, \emph{iii)} LD vocabularies formalized using the RDFS language may have less rich annotations compared to OWL ontologies (e.g., lack of synonyms and alternate labels), and \emph{iv)} LD sources, formalized using RDF and RDFS, may use \textit{reification} to capture information at a higher granularity. Hence, the methods to analyze semantic heterogeneity across biomedical ontologies cannot be translated ``as is'' to analyze semantic heterogeneity across biomedical linked data sources. 

A recent study found that the coverage of biomedical ontologies and public vocabularies is not sufficient to structure the entirety of biomedical linked data \cite{zaveri2016ontology}. Hence, linked data publishers end up creating their own custom vocabularies, and do not publish them in a standardized way for reuse in other projects. Recently, multiple studies have been conducted on the larger Linked Open Data cloud to investigate the availability and licensing of Semantic Web resources, as well as to provide a mechanistic definition of what constitutes a `link' in open knowledge graphs available on the Web \cite{haller2019links,polleres2019mored,debattistaevaluating}. However, these studies do not particularly focus on the semantic heterogeneity and vocabulary reuse across the Life Sciences Linked Open Data cloud \cite{hu2015link}

We use an Extraction Algorithm to extract and merge the schemas and vocabularies used by biomedical linked data sources for conducting our meta-analysis. The theory behind the Extraction Algorithm has identical characteristics to linked data profiling algorithms \cite{bohm2010profiling,hasnain2014roadmap,spahiu2016abstat,mihindukulasooriya2015loupe}. However, the linked data profiling algorithms have, in many cases, only been implemented over some of the popular SPARQL endpoints, do not account for variable SPARQL versions, require the retrieval of all instances and assertions that is often not scalable for LSLOD sources, or may not extend behind minimal schema extraction. Similarly, RDF graph pattern mining algorithms (e.g., evolutionary algorithms \cite{hees2016evolutionary}) often require exhaustive training samples of $(source, target)$ pairs which are often not possible to obtain for the LSLOD cloud.

\section*{Datasets}
\label{sec:alldatasets}
In the subsequent sections, we will provide you with details on the linked data graphs which were extracted from SPARQL endpoints exposed on the Life Sciences Linked Open Data (LSLOD) cloud. We also present details on additional corpora and datasets used as background knowledge sources to conduct our meta-analysis on the semantic heterogeneity across the linked data graphs.

\subsection*{Linked Data Graphs}
\label{sec:semhetld}

\begin{table*}[ht]
\centering
\begin{tabular}{l|p{9.5cm}}
\hline 
\textbf{Linked Data Project} & \textbf{Example RDF Graphs and Descriptions} \\
\hline
\hline
\textbf{Bio2RDF} \cite{callahan2013bio2rdf} & Several reprocessed data and knowledge sources such as:\\
& \textbf{DrugBank} \cite{wishart2006drugbank} --- Drug and drug target information \\
& \textbf{KEGG} \cite{kanehisa2000kegg} --- Biological pathways, proteins, and drugs \\ 
& \textbf{PharmGKB} \cite{hewett2002pharmgkb} --- Protein--drug--disease relations\\
& \textbf{CTD} \cite{davis2013comparative} --- Environmental chemical--protein interactions and pathway--disease relations \\
\hline
\textbf{EBI-RDF} \cite{jupp2014ebi} & Several proprietary data and knowledge sources such as: \\
& \textbf{Reactome} \cite{croft2014reactome} --- Biochemical reactions and pathways\\
& \textbf{BioSamples} \cite{barrett2012bioproject} --- Metadata about biological samples\\ 
& \textbf{ChEMBL} \cite{willighagen2013chembl} --- Drug-like molecules and activities \\
\hline
\textbf{UniProt} \cite{uniprot2008universal} & UniProt database on proteins and their characteristics \\
\hline
\textbf{MO-LD} \cite{deraspe2016making} & Model Organism databases on different species such as Fly, Mouse, Yeast, Rat, Human and Zebrafish  \\
\hline 
\textbf{NBDC RDF} \cite{kawashima2018nbdc} & Japanese Life Sciences databases on chemicals, structures, sequencing data, pathways, and diseases  \\
\hline
\textbf{Linked Life Data} \cite{linkedlifedatalink} & Several reprocessed data and knowledge sources such as: \\
& \textbf{EntrezGene} \cite{maglott2005entrez} --- Genes and variants in diseases \\ 
& \textbf{BioGrid} \cite{stark2006biogrid} --- Protein and chemical interactions\\
& \textbf{IntAct} \cite{kerrien2011intact} --- Protein--protein interactions  \\
\hline
\textbf{Linked TCGA} \cite{saleem2014big} & DNA methylation, gene expression and clinical data of cancer patients from The Cancer Genome Atlas \cite{weinstein2013cancer} \\
\hline
\textbf{NIH PubChem} \cite{fu2015pubchemrdf} & Databases on substances, compounds, structures, and biological assays stored in the PubChem repository\\
\hline
\textbf{WikiPathways} \cite{waagmeester2016using} & Database of biological pathways maintained using a crowd-sourcing architecture \\
\hline
\textbf{PathwayCommons} \cite{cerami2010pathway} & Data warehouse consisting of several biological pathways and molecular interactions datasets \\
\hline
\textbf{NLM MeSH} \cite{bushman2015transforming} & Medical Subject Headings used to index biomedical publications in MEDLINE repository \cite{medlinelink} \\
\hline
\textbf{DisGenet} \cite{pinero2016disgenet} & Data warehouse on genes and variants associated to diseases\\
\hline
\textbf{NextProt} \cite{lane2011nextprot} & Data warehouse on protein structures and interactions \\
\hline
\textbf{LinkedSPL} \cite{boyce2013dynamic} & Structured product labels of drugs represented as RDF\\
\hline
\hline
\end{tabular}
\caption{Linked data sources queried to generate the $\symG_{LSLOD}$ schema graph of the LSLOD cloud. }
\label{tab:ldsource}
\end{table*}

We extracted the schemas and vocabularies used in the RDF graphs exposed through $\approx 20$ SPARQL endpoints by different consortia. The entire list of the different linked data projects and details about sample RDF Graphs extracted are shown in \textbf{Table \ref{tab:ldsource}}. These RDF graphs include popular biomedical data sources such as DrugBank \cite{wishart2006drugbank}, Kyoto Encyclopedia of Genes and Genomes (KEGG) \cite{kanehisa2000kegg}, PharmGKB \cite{hewett2002pharmgkb}, Comparative Toxicogenomics Database (CTD) \cite{davis2013comparative}, exposed by the Bio2RDF consortium \cite{callahan2013bio2rdf}, UniProt \cite{uniprot2008universal}, ChEMBL \cite{willighagen2013chembl}, and Reactome \cite{croft2014reactome}, exposed by the EBI-RDF consortium \cite{jupp2014ebi}, and other sources such as The Cancer Genome Atlas (TCGA) \cite{weinstein2013cancer}, PubChem \cite{fu2015pubchemrdf}, and WikiPathways \cite{waagmeester2016using}. It should be noted that some of these projects (e.g., Bio2RDF \cite{callahan2013bio2rdf}) have reprocessed publicly available data and knowledge published by other groups (e.g., DrugBank \cite{wishart2006drugbank}), whereas in some cases data publishers themselves use Semantic Web technologies to provide access to their proprietary data as linked data (e.g., the European Bioinformatics Institute (EBI) RDF Platform \cite{jupp2014ebi}). 

These RDF graphs contain data and information on several biomedical entities such as drugs, genes, proteins, pathways, diseases, protein--protein interactions, drug--protein interactions, biological assays, and high-throughput sequencing. This is the largest collection of biomedical linked data sources that have functioning SPARQL endpoints or RDF dumps available on the Web. The endpoint URLs for these projects are listed at \url{http://onto-apps.stanford.edu/lslodminer}.

\subsection*{Corpus of Biomedical Ontologies}
\label{sec:ontodatasets}
We obtained a dump of 685 distinct biomedical ontologies stored in the BioPortal repository,\footnote{\url{https://bioportal.bioontology.org/}} the world's largest open repository of biomedical ontologies \cite{whetzel2011bioportal}. These ontologies were serialized using the N-triples format and versioned up to January 1, 2018. This dump did not contain some ontologies that were deprecated or merged with existing ontologies, or added to BioPortal after January 1, 2018. This corpus includes widely used ontologies such as Gene Ontology (GO) \cite{ashburner2000gene}, National Cancer Institute Thesaurus (NCIT) \cite{sioutos2007nci}, Chemical Entities of Biological Interest (ChEBI) Ontology \cite{hastings2013chebi}, and Systematized Nomenclature of Medicine - Clinical Terms (SNOMED CT) \cite{stearns2001snomed}, which are widely used in biomedicine for data annotation, knowledge management, and clinical decision support.

\subsection*{Corpus of Linked Open Vocabularies}
Linked Open Vocabularies\footnote{\url{https://lov.linkeddata.es/dataset/lov/}} (LOV) is a catalog of RDFS vocabularies available for reuse with the aim of describing data on the Web \cite{lov}. This catalog consists of standardized versions of general RDFS vocabularies that are often used to encode additional semantics for RDF graphs on the LSLOD cloud. Popular examples of vocabularies in this catalog include the W3C Provenance Interchange \cite{gil2013prov}, Simple Knowledge Organization System (SKOS) \cite{isaac2009skos}, and Schema.org \cite{guha2016schema}. We used 647 vocabularies that were present in the LOV catalog, as of January 1, 2018. The vocabularies in this catalog generally include concept descriptions (i.e., human-readable labels using the \texttt{rdfs:label} annotation properties).

\subsection*{Word Embedding Vectors Generated from the MEDLINE Corpus}
\label{sec:wordembed}
We generated 100-dimensional word embedding vectors from a corpus of approximately 30 million biomedical abstracts, stored in the MEDLINE repository \cite{medlinelink}, using the GloVe (Global Vectors for Word Representation) method \cite{pennington2014glove}. The steps taken to process biomedical abstracts (tokenization, medical entity normalization, etc.) were performed in accordance with those outlined in Percha, et al. \cite{percha2018global}. Word embeddings represent words as high-dimensional numerical vectors based on co-occurrence counts of those words as observed in the abstracts. These embedding vectors are available under a CC BY 4.0 license at FigShare,\footnote{\url{https://figshare.com/articles/Biomedical_
Word_Vectors/9598760.}} and have been used to map metadata fields in the BioSamples repository \cite{barrett2012bioproject} to ontology terms in the BioPortal repository \cite{gonccalves2019aligning}, as well as to identify biomedical relations in unstructured text \cite{kamdar2020text}.

\section*{Methods}
\label{sec:methods}
The meta-analysis conducted over the Life Sciences Linked Open Data (LSLOD) cloud relies on a set of extractors which navigate to each SPARQL endpoint that is available on the Web and that extract the schemas and vocabularies from those SPARQL endpoints. We have downloaded and stored RDF dumps in local SPARQL repositories to handle cases where an LSLOD source does not have a functional SPARQL endpoint but is available as an RDF data dump. The extractors use an Extraction Algorithm, whose technical underpinnings are provided in \textbf{Supplementary Material}.

\subsection*{Extraction Algorithm}
\label{meth:extract}
\begin{figure}[!htb]
\begin{centering}
\includegraphics[scale=0.35]{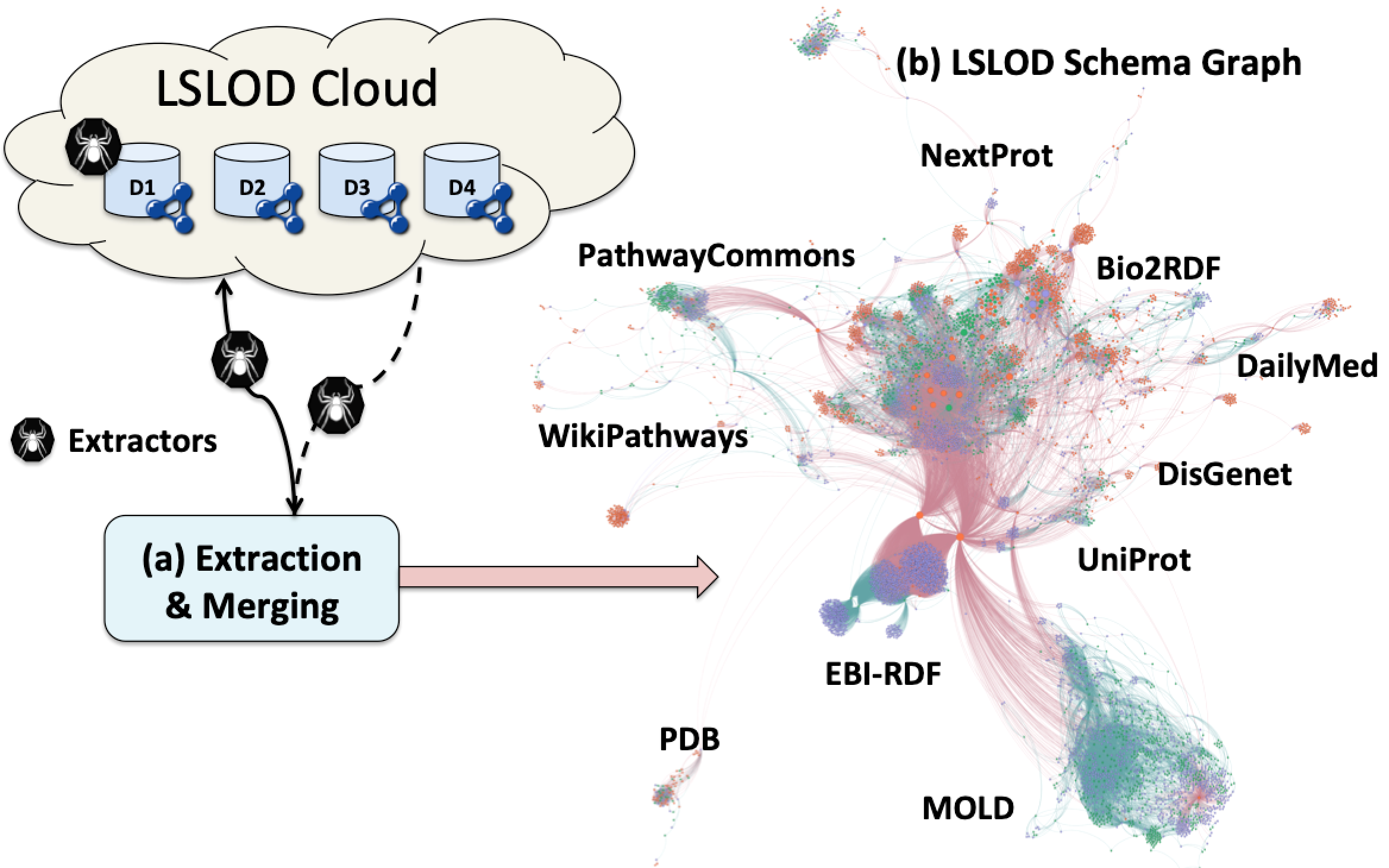}
\par\end{centering}
\caption[Diagrammatic Representation of the Extraction Algorithm]{\textbf{Diagrammatic Representation of the Extraction Algorithm.} During the extraction phase, automated extractors (represented as black spiders here) query each SPARQL endpoint in the LSLOD cloud and extract the schemas (i.e., classes, object properties, data properties, URI patterns, graph patterns) to generate the LSLOD Schema Graph $\symG_{LSLOD}$. A portion of $\symG_{LSLOD}$ is visualized here, using a force-directed network, with classes (purple nodes), object properties (green nodes) and data properties (orange nodes) extracted from multiple sources, such as Bio2RDF, EBI-RDF, etc.}
\label{fig:patternminer}
\end{figure}

The Extraction Algorithm uses a set of SPARQL query templates (listed in \textbf{Supplementary Material}) to extract schemas (e.g., classes, properties, domains, ranges) as well as sample instances of those classes and their property values from the SPARQL endpoints in the LSLOD cloud. The input to the Extraction Algorithm is the SPARQL endpoint, whose version of SPARQL can be automatically detected based on the SPARQL queries and keywords supported. The Extraction algorithm not only accounts for different versions of SPARQL supported at the remote SPARQL endpoints, but also uses alternative SPARQL queries if the remote endpoint times out during the extraction phase. 

An extractor that encapsulates the Extraction Algorithm selects the set of SPARQL query templates for execution against the input SPARQL endpoint, given its version. The extractor queries the SPARQL endpoint and extracts the RDF graphs exposed through the SPARQL endpoint. For each RDF graph, the extractor extracts the set of classes and the total number of instances for each class. The extractor identifies the properties for which a specific class serves as the domain for all assertions, and also retrieves the range classes for object properties, or the datatypes for data properties. The Extraction Algorithm also extracts a random sample (without replacement) of 2,000 instances for each class and 2,000 property assertions for a class--property combination, and generates summary characteristics (e.g., categorical, datatype, namespace, median length, regular expression patterns) of the associated data. 

Classes, properties, and datatypes, are represented as nodes in the schema graph $\symG_{{\symT}-Box}$ extracted from the SPARQL endpoint by the specific extractor. The edges link property nodes to class nodes and datatype nodes depending on the domain and range descriptions of the corresponding property. The nodes and edges are annotated by the count and summary characteristics of corresponding instances or property assertions respectively. This schema graph is a fragment of the actual $\symT$-Box of the RDF graph. The $\symG_{{\symT}-Box}$ graphs extracted for all RDF graphs in the LSLOD cloud are merged to generate the schema graph $\symG_{LSLOD}$ of the entire LSLOD cloud. A diagrammatic representation of this method and a subset of the merged $\symG_{LSLOD}$ schema graph are shown in \textbf{Figure \ref{fig:patternminer}}.

The Extraction Algorithm and the SPARQL query templates used by the algorithm are documented in \textbf{Supplementary Material} in greater technical detail.

\subsection*{Determining Vocabulary Reuse across Biomedical Linked Data Sources}
\label{meth:vocabreuse}
Estimating the extent of observed vocabulary reuse across different biomedical linked data sources requires identifying the classes and properties most commonly reused from standard ontologies and vocabularies. We use the corpus of biomedical ontologies stored in the BioPortal repository \cite{whetzel2011bioportal} as well as popular RDFS vocabularies published on the Linked Open Vocabularies repository \cite{lov} for this goal. We also use the $\symG_{LSLOD}$ schema graph extracted from the LSLOD cloud. 

For each URI in the $\symG_{LSLOD}$ schema graph (class, property, or sample instance URI), the \textit{origin} of the URI is determined using a heuristic approach \cite{kamdar2015investigating} by converting each URI to lowercase and using \emph{regular expression} filters constructed from namespaces (e.g. \texttt{ncicb.nci.nih.gov} for NCIT \cite{sioutos2007nci}) and common identifier patterns. 

For each linked data source, we compute the following statistics:
\begin{enumerate}[noitemsep]
\item The number of ontologies and vocabularies reused in the source
\item The percentage of schema elements reused from external LSLOD sources
\item The percentage and distribution of classes whose entities are mapped or linked to other entities in external LSLOD sources (interlinking)
\item The percentage and distribution of classes whose entities are mapped or linked to entities of other classes in the same LSLOD source (intralinking)
\item The percentage of entities reused or mapped to external LSLOD sources
\end{enumerate}

Previously, we outlined a method to estimate reuse of classes across multiple biomedical ontologies \cite{kamdar2017systematic}. We repurposed this method to estimate vocabulary reuse across biomedical linked data sources. We generated an undirected network composed of different URIs from the $\symG_{LSLOD}$ schema graph as nodes (i.e., the classes and the properties), and the edges indicating reuse and mapping between the different schema element URIs. ``Patterns'' of instance URIs are also represented as nodes in this network to check if instances are reused or mapped to classes in biomedical ontologies (i.e., semantic mismatch). We extract connected components from this network and use \textbf{Equation \ref{eqn:reuse}} to estimate vocabulary reuse.

\begin{equation}
\label{eqn:reuse}
Vocabulary\;Reuse = \frac{\sum_{j|T_j \in \symM_r}n_j-k}{N}
\end{equation}

In the above equation, assume that the network is composed of $k$ connected components $\{\symT_0, \symT_1, \ldots, \symT_k\}$, and each component $\symT_j$ is formed from $n_j$ schema elements (i.e., $\{t_{0j},t_{1j},\ldots,t_{nj}\}\in \symT_j$). Assume that $N$ is the total number of schema element URIs in the network. The number of terms in a component $T_j$ must follow $1 < n_j < N$ (i.e., components with a single term are not allowed). All schema elements in one component exhibit some reuse with respect to each other.


\subsection*{Detecting Communities of Related Content in the LSLOD Cloud}
\label{meth:labelmismatch}

\begin{figure*}[!htb]
\begin{centering}
\includegraphics[scale=0.23]{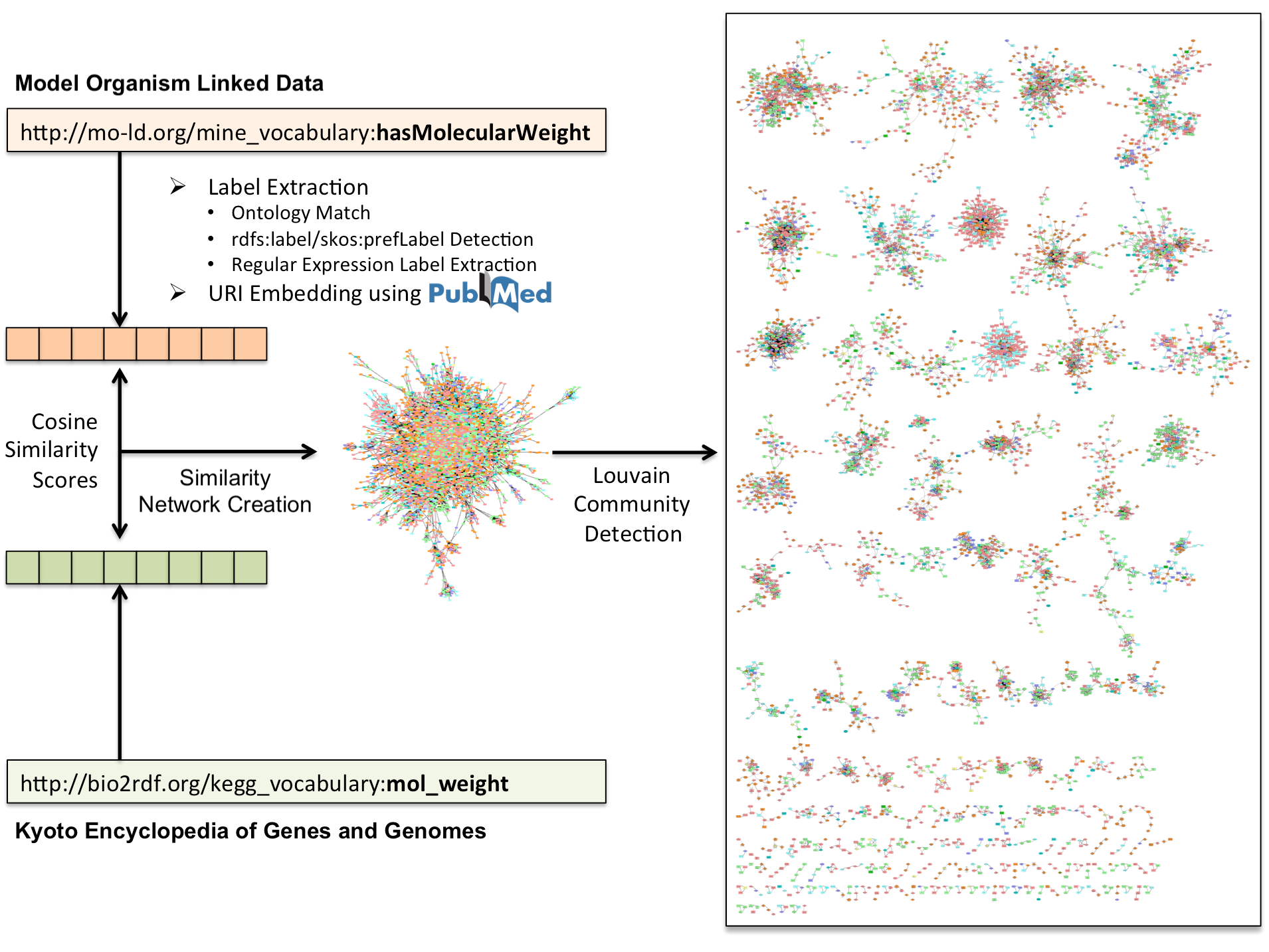}
\par\end{centering}
\vspace{-2mm}
\caption[Diagrammatic representation of the Algorithm to Detect Communities of Similar Content across LSLOD Cloud]{\textbf{Diagrammatic representation of the Algorithm to Detect Communities of Similar Content across LSLOD Cloud:} The labels for two URIs are extracted either \emph{i)} by matching to a reused ontology, or \emph{ii)} by using annotation properties such as \textit{rdfs:label} or \textit{skos:prefLabel}, or \emph{iii)} by using regular expressions on the URI. URI embeddings are generated from the extracted labels, by using a background source of MEDLINE biomedical abstracts. A similarity network is created with nodes representing the different URIs from the $\symG_{LSLOD}$ schema graph and edges representing the cosine similarity scores between two URI embeddings. We detect communities in this network using the Louvain method. The trivial example shown here is related to similar ``molecular weight'' URIs between the Model Organism - linked data (MO-LD) project and the Bio2RDF KEGG linked data graph.}
\label{fig:labelmismatch}
\end{figure*}

To detect clusters of related content in the LSLOD cloud, we used a 5-step algorithm that maps similar schema elements (e.g., classes and properties) between two different sources in the LSLOD cloud. A diagrammatic representation of this algorithm is shown in \textbf{Figure \ref{fig:labelmismatch}}. This algorithm uses the schema elements extracted from the various LSLOD public SPARQL endpoints (see \textbf{\ref{tab:ldsource}}) and included in the $\symG_{LSLOD}$ graph. These schema elements may or may not be reused from a common vocabulary or ontology (vocabulary reuse in LSLOD cloud was determined using the methods described in the previous section). The algorithm uses two sources of background knowledge to generate the mappings between two different schema elements: \emph{i)} word embeddings generated from the MEDLINE corpus, and \emph{ii)} the set of biomedical ontologies and vocabularies that are reused across biomedical LD sources.

\subsubsection*{Label Extraction}

In the first sweep, the algorithm extracts a label for each schema element URI (i.e., class, object property or data property) in the $\symG_{LSLOD}$ graph. This is done through three approaches in a sequential flow:
\begin{enumerate}[noitemsep]
    \item A label is extracted from the values for the \texttt{rdfs:label} \cite{mcbride2004resource} or \texttt{skos:prefLabel} \cite{isaac2009skos} annotation properties, if the $\symG_{LSLOD}$ schema element has already been reused from existing biomedical ontologies and vocabularies (see \textbf{Datasets}).
    \item A label is extracted by querying the source (i.e., the SPARQL endpoint) of the schema element for \texttt{rdfs:label} or \texttt{skos:prefLabel} annotation assertions.
    \item A label is generated using Regular Expression (\textit{RegExp}) filters from the URI of the schema element. Specifically, the filters deal with cases where separators (e.g., "-", "\_") and camel case (e.g. hasMolecularWeight) is present in the URI. In \textbf{Figure \ref{fig:labelmismatch}}, the labels ``Has Molecular Weight'' and ``Mol Weight'' will be extracted from the two URIs respectively. 
\end{enumerate}
   
\subsubsection*{URI Embedding Generation} 
Each schema element URI is then represented in a high dimensional space using word embedding vectors. Specifically, a vocabulary of words\footnote{A word should appear in at least five separate abstracts for inclusion in the vocabulary.} is generated from 30 million PubMed biomedical publication abstracts. A URI embedding vector is generated by computing a weighted average of the embedding vectors of the words in the URI label, with the weights being the Inverse-Document-Frequency (IDF) statistic for each word. A default embedding vector and IDF statistic is created for any word that is not present in the vocabulary. The equation to generate a URI embedding vector is shown below. Here, $\mathbf{x}_{(w_i)}$ represents the 100-dimensional word embedding vector, and $\symL(URI)$ is the URI label.
        \begin{align}
            \mathbf{x}_{(URI)} = \frac{\sum_{w_i \in \symL(URI)}idf_{(w_i)}*\mathbf{x}_{(w_i)}}{\sum_{w_i \in \symL(URI)}idf_{(w_i)}} 
        \end{align}

\subsubsection*{Cosine Similarity Score Computation} 

For any two schema element URIs that are not present in the same source, cosine similarity scores are computed using their URI embedding vectors. The equation to compute the similarity between two elements is presented below.
        \begin{align}
            Sim(A, B) = \frac{\mathbf{x}_{(URI_A)}\cdot\mathbf{x}_{(URI_B)}}{||\mathbf{x}_{(URI_A)}||||\mathbf{x}_{(URI_B)}||} = \frac{\sum_{i=1}^{n=100}x_{(URI_A)i}*x_{(URI_B)i}}{\sqrt{\sum_{i=1}^{n=100}x_{(URI_A)i}^2}\sqrt{\sum_{i=1}^{n=100}x_{(URI_B)i}^2}}
        \end{align}

\subsubsection*{Creating a Similarity Network}

A similarity network is created where the different schema element URIs represent the nodes of this network. An edge exists between two nodes if: \emph{i)} the cosine similarity score between the corresponding URIs is greater than the threshold of 0.75, and \emph{ii)} there exists a one-to-one mapping between the two schema elements for the particular combination of linked data sources. 

For example, consider three URI nodes $n_1, n_2, n_3$, such that, $n_1, n_2 \in \symLD_1$ and $n_3 \in \symLD_2$. If $Sim(n_1, n_3) = 0.87, Sim(n_2, n_3) = 0.93$ then an edge will only be created between $n_2$ and $n_3$ nodes. This criteria has an underlying assumption that there should exist a unique alignment between similar schema elements in different linked data sources. While the cosine similarity scores in the previous step can, by itself, aid in detection of similar schema elements (e.g., all schema elements pertaining to \textit{Molecular weight}), this similarity network will aid in the detection of sources that contain relevant information pertaining to a given biomedical concept or relation type.

\subsubsection*{Community Detection using Louvain Method} 

Finally, the Louvain method \cite{blondel2008fast} for community detection in weighted networks is used. The cosine similarity scores form the edge weights in the similarity network. This method has been used to detect communities of users using social networks and mobile phone usage data, as well as detect species in network-based dynamic models \cite{clauset2004finding, markovitch2018predicting, de2011generalized}. The Louvain method maximizes a `modularity' statistic $Q$, which measures the density of edges inside communities compared to edges between communities. In the equation below (reproduced from Blondel, et al. \cite{blondel2008fast}), $2m$ is the sum of all edge weights in the graph (i.e., $m=\frac{1}{2}\sum_{ij}Sim(i,j)$) and $k_i$ is the sum of all weights to edges attached to node $i$ (i.e. $k_i = \sum_{j}Sim(i,j)$). $\delta(c_i, c_j)$ is a delta function that looks at the community assignments ($c_i$ and $c_j$) for connected nodes $i$ and $j$.
    \begin{align}
        Q &= \frac{1}{2m}\sum_{ij}\Big[Sim(i,j) - \frac{k_ik_j}{2m}\Big]\delta(c_i, c_j) \\
        \delta(c_i, c_j) &= \left\{
        	\begin{array}{ll}
        	1 & \mbox{if } (c_i = c_j) \\
        	0 & \mbox{if } (c_i \neq c_j)
        	\end{array}
        \right. 
    \end{align}

The communities are visualized using the Cytoscape platform \cite{shannon2003cytoscape}.

\section*{Results}

\label{sec:results}
\subsection*{LSLOD Schema Graph $\symG_{LSLOD}$}
\label{data:lslodsources}
Using the Extraction Algorithm, we extracted $99$ RDF Graphs\footnote{This number varies since a database may divided into multiple graphs with different URIs to be stored in a SPARQL endpoint. We have merged similar RDF graph URIs based on the actual data source.} from $\approx20$ SPARQL endpoints. The total number of classes, object properties, data properties, and datatypes extracted and linked in the $\symG_{LSLOD}$ schema graph were $57,748$, $4,397$, $8,447$ and $24$ respectively. The extractors can function on any SPARQL endpoint irrespective of its \emph{i)} version, \emph{ii)} support for different SPARQL keywords \cite{prud2008sparql}, and \emph{iii)} the size of the underlying RDF graphs. The Extraction Algorithm can generally process most SPARQL endpoints within 4--8 hours. A visualization of a portion of the extracted $\symG_{LSLOD}$ schema graph is shown in \textbf{Figure \ref{fig:patternminer}B}. A snippet of the extracted information for a given source is shown in \textbf{Listing \ref{lst:extractor}}. 

Through an empirical analysis, we have found some overlap between the different schema elements, based on the way linked data publishers model the underlying data (i.e., some object properties in a given source are used as data properties in another linked data source). We found standard datatypes defined under the XML Schema Definition (XSD) \cite{biron2004xml} and RDFS \cite{mcbride2004resource} such as \texttt{xsd:string}, \texttt{xsd:integer}, \texttt{xsd:float}, \texttt{xsd:dateTime}, \texttt{xsd:boolean}, \texttt{rdfs:Literal}. The analysis also found an ontology class \texttt{UO:0000034} (Units Ontology \cite{gkoutos2012units} \textsc{Week} class) used as a datatype. In some cases, we also found that RDF graphs may exhibit Semantic Mismatch where instances are aligned to classes from an exhaustive ontology, such as ChEBI \cite{hastings2013chebi} or NCIT \cite{sioutos2007nci} that have more than $ 50,000$ classes, using the \texttt{rdf:type} property. \textit{Semantic Mismatch} in linked data is a form of semantic heterogeneity. Since \texttt{rdf:type} is a predicate in almost all the SPARQL query templates used by the Extraction Algorithm, semantic mismatch significantly increases the number of classes in the $\symG_{\symT-Box}$, and consequently the number of queries that need to be formulated for each class significantly increases (See \textbf{Supplementary Material}).

\begin{lstlisting}[captionpos=b, caption={Information from the Bio2RDF DrugBank RDF Graph extracted by the Extraction Algorithm}, label=lst:extractor, basicstyle=\ttfamily,frame=single, language=Java, float,floatplacement=H]
Source: DrugBank
Source URI: http://bio2rdf.org/drugbank_resource:bio2rdf.dataset.drugbank.R3 
Classes:    
    Class: Drug
        Class URI: http://bio2rdf.org/drugbank_vocabulary:Drug
        Sample Instances: ['http://bio2rdf.org/drugbank:DB03536', ...]
    Class: Drug-Drug-Interaction
        Class URI: http://bio2rdf.org/drugbank_vocabulary:Drug-Drug-Interaction
        ...
Properties: 
    Object Property: target
        Property URI: http://bio2rdf.org/drugbank_vocabulary:target
        Property Realization: drugbank:Drug -> drugbank:target
            Domain: Drug, Range: Target
            Sample Assertion Values: ['http://bio2rdf.org/drugbank:BE0000059', ...]
    Data Property: molecular-weight
        Property URI: http://bio2rdf.org/drugbank_vocabulary:molecular-weight
        Property Realization: drugbank:Enzyme -> drugbank:molecular-weight
            Domain: Enzyme, Range: Float
            Sample Assertion Values: [52221.099999999999, 56345.0, 57530.0, ...]
\end{lstlisting}

\subsection*{Vocabulary Reuse across Linked Data Sources}
\begin{figure*}[!htb]
\begin{centering}
\includegraphics[scale=0.38]{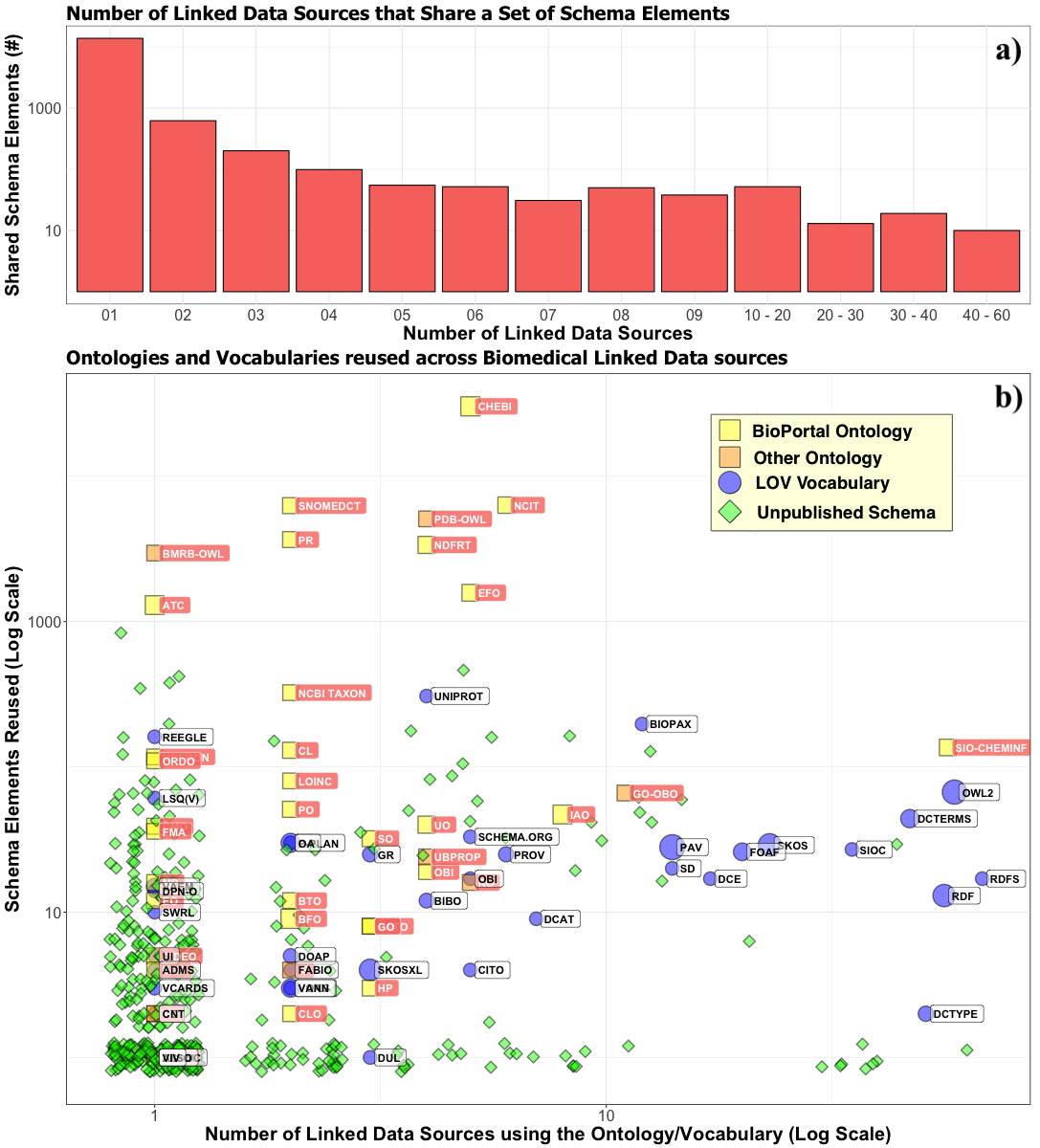}
\par\end{centering}
\caption[Vocabulary reuse across biomedical linked data sources]{\textbf{Vocabulary reuse across biomedical linked data sources (RDF graphs): a)} A histogram of schema elements that are shared across multiple LD Sources. It can be seen that most LD Sources use a unique schema or vocabulary whose elements are not reused across any other data source. However, there are at least 100 schema elements that are shared by $>10$ LD Sources. \textbf{b)} The different vocabularies (published at LOV repository) and ontologies (BioPortal and other OWL ontologies) whose elements are reused in the schemas of LD sources. Several LD sources either use unpublished schemas or use elements from common vocabularies (e.g. DCTerms and SKOS).}
\label{fig:VocabularyReuse}
\end{figure*}
We analyzed the $70,592$ total schema elements (classes, object properties, and data properties) extracted from the linked data sources in $\symG_{LSLOD}$ for vocabulary reuse. In this research, we looked only at reuse of schema elements from vocabularies published at the Linked Open Vocabularies (LOV) repository \cite{lov} and ontologies that are either published at the BioPortal repository \cite{whetzel2011bioportal} or on the Web as OWL files available for download (and have \texttt{.owl} in the namespace prefix, for easy search). 

\textbf{Figure \ref{fig:VocabularyReuse}} shows the results of the analyses. \textbf{Figure \ref{fig:VocabularyReuse}a} depicts a histogram of schema elements that are shared across $X$ number of linked data sources. It can be observed that most LD sources are composed of unique schema elements that are present only in that particular source (i.e., the first histogram bar from the left). However, there are at least $>100$ schema elements that are shared between $>10$ LD sources (i.e., the last four histograms from the right). 

Terms are reused in biomedical linked data graphs from 34 BioPortal ontologies, 7 other biomedical ontologies, available on the Web but not stored in the BioPortal repository, and 43 LOV vocabularies (\textbf{Figure \ref{fig:VocabularyReuse}b}). Some of the reused ontologies and vocabularies are very popular in biomedicine (e.g., ChEBI \cite{hastings2013chebi}, SNOMED CT \cite{stearns2001snomed}, NCIT --- National Cancer Institute Thesaurus \cite{sioutos2007nci}, GO --- Gene Ontology \cite{ashburner2000gene}) and in the Semantic Web community (e.g., DCTerms Metadata Description Model \cite{kunze2007dublin}, RDFS \cite{mcbride2004resource}, SKOS --- Simple Knowledge Organization System \cite{isaac2009skos}, Data Catalog Vocabulary \cite{maali2014data}) respectively. 

However, most of these biomedical ontologies are used only in the schemas of a few LD sources (X-axis), and moreover only a few elements from these ontologies are reused (Y-axis). For example, the Anatomical Therapeutic Chemical Classification terminology \cite{skrbo2004classification} --- popularly used to classify drugs according to their mechanism of action --- is used in the schema of only one LD source.  It can be seen that several LD sources use unpublished schemas (green diamonds), hence, a large number of unique schema elements are present in only one source in \textbf{Figure \ref{fig:VocabularyReuse}a}. 

Several LD sources do reuse schema elements from a set of widely used vocabularies ($\approx 100$ as per \textbf{Figure \ref{fig:VocabularyReuse}a}). However, the elements of these vocabularies are not useful for data integration and integrated querying from a biomedicine perspective. It should be noted that several elements from the combination of the Semantic Science Ontology and Chemical Information Ontology (SIO-CHEMINF) \cite{dumontier2014semanticscience} are reused across several LD sources and may be a popular resource for the emergence of a common data model for integrated querying of these sources\footnote{A large number of these LD sources however only use one particular element from SIO ontology.}. 

Using \textbf{Equation \ref{eqn:reuse}}, vocabulary reuse was determined to be $19.86\%$, which may indicate favorable reuse across LD sources in comparison to ontologies. However, it should be noted that this statistic may be driven by the reuse of 30,251 ChEBI classes in a few sources. We also determined several cases of \textit{semantic mismatch} where classes from popular biomedical ontologies in the BioPortal repository (e.g., SNOMED CT, NCIT, GO) are mapped to instances in these LD sources. 

\begin{figure*}[!htb]
\begin{centering}
\frame{\includegraphics[scale=0.11]{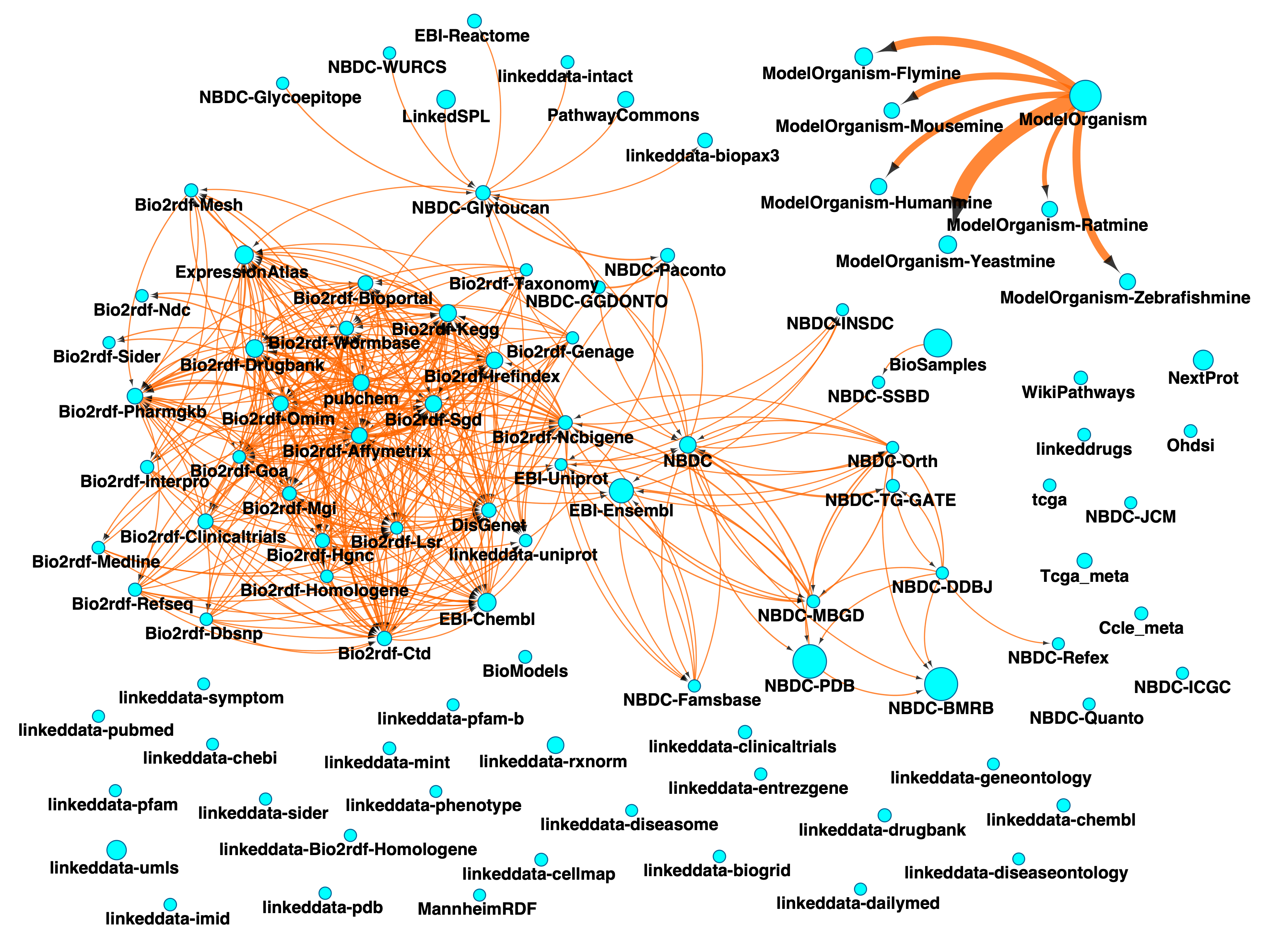}}
\par\end{centering}
\caption[Interlinking and intra-linking across biomedical linked data sources]{\textbf{Interlinking and intra-linking across biomedical linked data sources (RDF graphs).} Each unique RDF Graph retrieved by the Extraction Algorithm is shown as a node in this network. The size of the node indicates the number of unique object properties used to link entities in different classes from the same source. The size of the edge indicates the number of unique object properties used to link entities in different classes in different sources.}
\label{fig:interlinkLD}
\end{figure*}

\textbf{Figure \ref{fig:interlinkLD}} shows the different RDF graphs extracted as nodes in a network. The size of the node indicates the number of unique object properties used to link entities in different classes from the same source (Intra-Linking). The size of the edge connecting two different nodes indicates the number of unique object properties used to link entities in different classes in different sources or RDF Graphs (represented using the corresponding node). This network diagram can be perceived to be similar to the LSLOD cloud diagram\footnote{\url{http://lod-cloud.net}}. It can be observed that the LSLOD cloud is not actually as densely connected as is often indicated through the LSLOD cloud diagram. Several sources, which may exhibit a higher degree of intra-linking are not inter-linked with other RDF graphs and exist as stand-alone data sources converted using RDF. The highly appreciated efforts by the Bio2RDF \cite{callahan2013bio2rdf} and EBI-RDF \cite{jupp2014ebi} consortia can be seen in the dense interlinking between the consortia-published sources. 

\subsection*{Communities of Related Content in the LSLOD Cloud}
While the low level of vocabulary reuse across linked data sources present a bleak picture of the Semantic Web vision toward facilitating Web-scale computation and integrated querying, it should be noted that there is significant similarity of content across these sources. 

We generated 100-dimensional word embedding vectors and inverse document frequency (IDF) statistics for a vocabulary of 2,531,989 words from approximately 30 million MEDLINE biomedical publication abstracts. Using these word embedding vectors and IDF statistics, we generated 100-dimensional URI embedding vectors for different schema elements URIs in the $\symG_{LSLOD}$ schema graph. Whereas the word tokens for most URI labels were found in the vocabulary, 3,340 word tokens, such as `differn', `heteronucl', and `neoplas' were not found either due to typos or due to minimal occurrence frequency in publications. We represented these words using the default embedding vector and IDF statistic. We computed cosine similarity scores to detect additional mappings between different schema elements. For example, the \texttt{drugbank:Molecular-Weight} class has higher cosine similarity scores to related schema elements, such as \texttt{kegg:mol\_weight}, \texttt{biopax:molecularWeight}, and \texttt{MOLD:hasMolecularWeight} data properties, as well as the \texttt{CHEMINF:000198} (\textsc{Molecular Weight Calculated by Pipeline Pilot}) class.

A similarity network was created with 12,613 schema elements as nodes in this network. This number of nodes is less than the total number of schema elements in the $\symG_{LSLOD}$ schema graph (70,587). We  excluded domain-specific classes that were reused from ChEBI and other ontologies through semantic mismatch. We also excluded URIs for which labels could not be extracted. 

Edges were created between these nodes using the cosine similarity scores between the corresponding schema elements. There were 169,005 pairs of schema element URIs that satisfied the first criterion of having a cosine similarity value above the threshold of 0.75. This threshold was predetermined after an exploratory analysis of pairs of schema elements that satisfied this criterion. While it would make sense to have a higher threshold ($\approx0.95$) for a higher degree of certainty on similarity, having a threshold of 0.75 allowed the inclusion of pairs, such as \texttt{obo:regulates} $\leftrightarrow$ \texttt{drugbank:mechanism-of-action}: 0.75, \texttt{chembl:mechanismDescription} $\leftrightarrow$ \texttt{drugbank:mechanism-of-action}: 0.85, \texttt{faldo:location} $\leftrightarrow$ \texttt{drugbank:cellular-location}: 0.89. After applying the second criterion of one-to-one mapping between two schema elements for a particular combination of linked data sources, the network had 22,855 pairs of schema-element URIs for which edges were created. 

\label{res:labelmismatch}
\begin{figure*}[t!]
\begin{centering}
\frame{\includegraphics[scale=0.125]{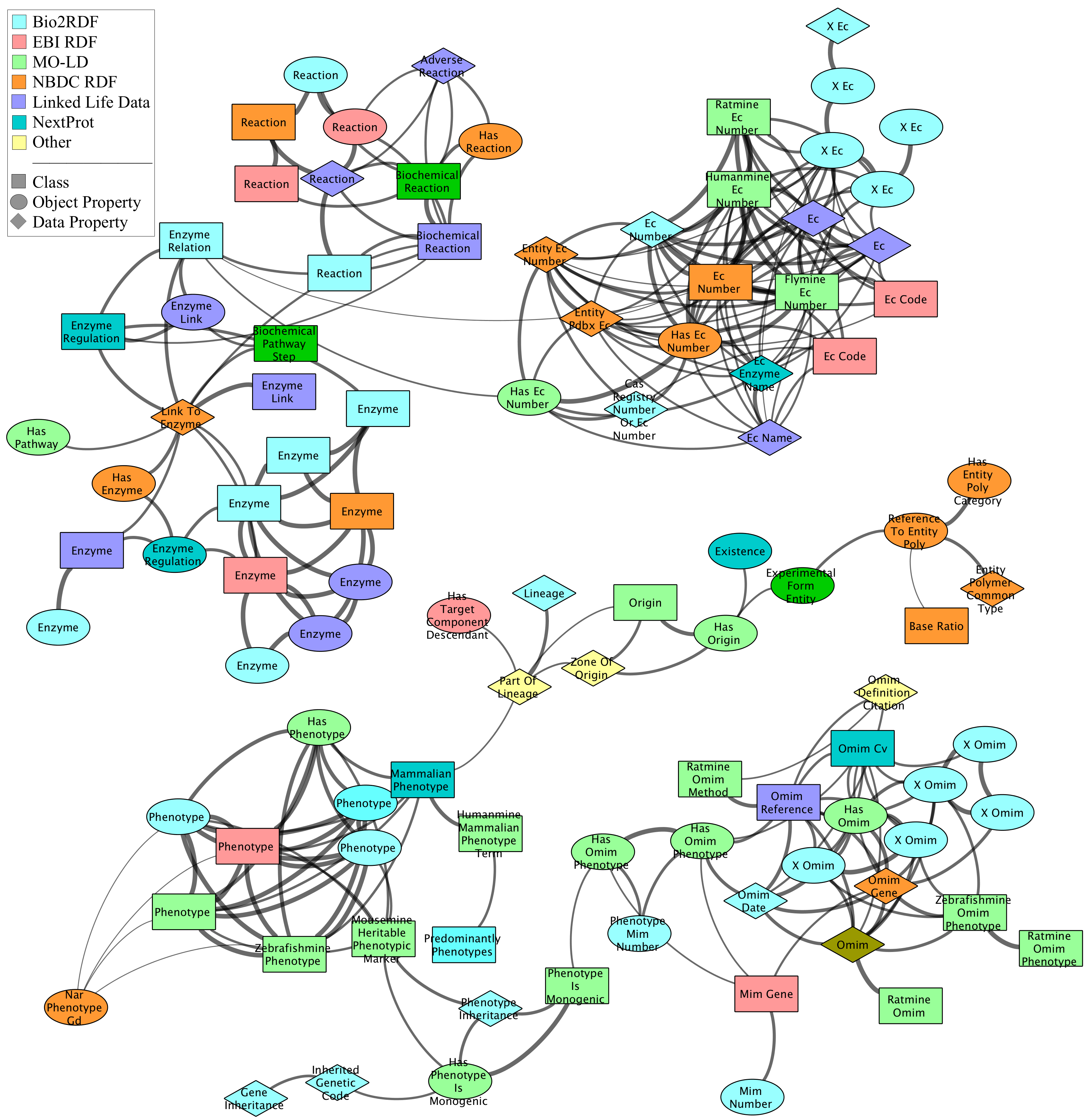}}
\par\end{centering}
\caption[Communities of Similar Content across Biomedical Linked Data]{\textbf{Communities of Similar Content across Biomedical Linked Data.} Two communities composed of schema elements related to \textsc{Enzyme} and \textsc{Phenotype} concepts are visualized in a greater detail. The nodes indicate a distinct schema element (distinct URI), and are colored according to the linked data project they are present in. The shape of the nodes indicates the type of schema element (class, object property, data property). The width of an edge between two nodes is proportional to the cosine similarity score between the URI embeddings of the corresponding schema element. Smaller sub-communities related to \textsc{Reaction}, \textit{Enzyme Commission Number} and \textit{OMIM Phenotype Identifier} (Online Mendelian Inheritance in Man) are also observed.}
\label{fig:labelmismatchcomm}
\end{figure*}

We used the Louvain method for detecting communities in the weighted similarity network. The method terminated after achieving a maximum modularity statistic (i.e., the density of edges inside communities compared to edges between communities) of 0.817296. We detected 6,641 communities. While most of these communities had minimal membership (i.e., $\leq10$ schema elements were members), there were 51 communities with more than $10$ schema elements as members. On visual inspection using the Cytoscape network visualization platform, we determined that these communities primarily consisted of schema elements related to a few key biomedical concepts or relation types. These communities are partially visualized in \textbf{Figure \ref{fig:labelmismatch}}, with two communities composed of schema elements related to \textsc{Enzyme} and \textsc{Phenotype} concepts visualized in a greater detail in \textbf{Figure \ref{fig:labelmismatchcomm}}. \textsc{Enzyme} entities are specialized proteins which act as biological catalysts to accelerate biochemical reactions, whereas \textsc{Phenotype} entities are composite observable characteristics or traits of an organism. 

In \textbf{Figure \ref{fig:labelmismatchcomm}}, each node is a distinct schema element URI (i.e., class, object property or data property) observed in the $\symG_{LSLOD}$ LSLOD schema graph. These nodes are colored based on the linked data project they are primarily present in (e.g., Bio2RDF, EBI-RDF). It should be noted that each linked data project (e.g., Bio2RDF) may publish multiple sources as distinct RDF graphs (e.g., KEGG \cite{kanehisa2000kegg}, DrugBank \cite{wishart2006drugbank}). The shape of the nodes is reminiscent of the type of schema element --- classes are represented as rectangular nodes, object properties as elliptical nodes, and data properties as diamond nodes. The nodes are labeled using the extracted labels, but the underlying URIs may be completely different. The width of the edge connecting different nodes is proportional to the cosine similarity scores between the URI embeddings of the connected schema elements.

It can easily be seen that different schema element URIs may be used across different sources (even if the sources are published by the same linked data project) to model and represent similar content and information (e.g., information on enzymes or phenotypes). It also showcases that data and knowledge relevant to a particular biomedical entity may be scattered across multiple sources (i.e., knowledge on a particular \textsc{Enzyme} entity can be retrieved from Bio2RDF, EBI-RDF and NextProt, and in many cases this knowledge may be novel). Other communities composed of schema elements focused on other key biomedical concept or relation types are also observed (e.g., \textsc{Drug}, \textsc{Publication}, \textsc{Chromosome}). The primary composition of a few of these larger communities ($>30$ members)  is listed in \textbf{Table \ref{tab:communitycompo}}. It should be noted that these communities may be composed of multiple sub-communities. For example, a sub-community of schema elements related to the \textsc{Reaction} concept is linked to the main cluster of nodes related to the \textsc{Enzyme} concept, since \textsc{Enzyme} entities are often associated with \textsc{Reaction} entities.

\begin{table*}[]
\centering
\begin{tabular}{c|c|p{10cm}}
\hline
\textbf{Com.} & \textbf{Total} & \textbf{Key Biomedical Concepts}  \\
\textbf{Id.} & \textbf{Elements} & \\
\hline
\hline
5023 & 549 & \textbf{Clinical} (Age, Treatment, Summary, Dosage, Diagnosis) \\
& & \textbf{Cancer} (Stage, Tumor, Cell, Sample, Rate) \\
& & \textbf{Genomics} (Experiment, Target, Function, Interaction, Correlation, Assembly, Association, Expression, Genetic)\\
\hline
3098 & 281 & \textbf{Protein and Genomic Sequences} (Sequence, Position, Start, Length, Protein, End, Repeat, Gene, Match, Begin, Translation, Exact, Exon, Frame, Distance) \\
\hline
2842 & 271 & \textbf{Clinical} (Date, Condition, Entry, Process, Person, Diagnosis, Participant, Outcome, Indication, Therapy, Evidence, Access) \\
\hline
4346 & 260 & \textbf{Data} (Code, Reference, Database, Sample, Data, Ontology, Semantic, Experiment, Dataset, Assembly, Registry)\\
\hline
2879 & 231 & \textbf{Clinical} (Clinical, Group, ICD) \\
&  & \textbf{Genetic} (Gene, Entity, Ncbigene, Species, Atom, Ref) \\
\hline
3945 & 228 & \textbf{Biomedical Relations} (Disease, Relationship, Expression, Gene, Level, Sample, Drug, Interaction, Component) \\
\hline
2515 & 211 & \textbf{Protein} (Protein, Site, Region, Molecule, Domain, Binding, Interaction, Active, Atom, Canonical, Primary, Structure) \\
\hline
6376 & 210 & \textbf{Literature} (Description, Author, Entity, Details, Note, Article, Issue, List, Type, Study, Entry, Document, Data, Address) \\
\hline
1417 & 192 & \textbf{Identifiers} (Uniprot, Identifier, Accession, Ensembl, Taxonomy, PDB, Genbank, Gene, MGI, Database, Id, Refseq, NCBI, CDS) \\
\hline
1632 & 171 & \textbf{Identifiers} (Id, Taxon, Mesh, Entity, Type) \\
& & \textbf{Genomics} (Platform, Entry, Sample, Genus, Array, Gene, List) \\
\hline
3249 & 171 & \textbf{Genotype--Phenotype Associations} (Gene, Symbol, Association, Product, Variant, Tag, Locus, Phenotype, Allele, Disease) \\
\hline
5652 & 158 & \textbf{Genomics} (RNA, Probe, Strand, Direction, Region, DNA) \\
\hline
1446 & 132 & \textbf{Software} (Version, Software, File, Format, Image, Email, Scale) \\
\hline
2535 & 126 & \textbf{Drug} (Drug, Chemical, Action, Ingredient, Bond, Activity) \\
\hline
2421 & 110 & \textbf{Chromosome} (Location, Chromosome, Entity, Cellular) \\
\hline
2317 & 105 &  \textbf{Pathway} (Pathway, Molecular, Formula) \\
& & \textbf{Unit} (Weight, Body, Units, Volume, Unit)\\
\hline
1806 & 101 & \textbf{Source} (Source, Natural, Vector, Derived, Tissue) \\
\hline
2672 & 99 & \textbf{Literature} (Count, Text, Comment, Editor, List, Data, Notes) \\
\hline
6316 & 97 & \textbf{Ontology} (Annotation, Ontology, Type, Structure) \\
\hline
5969 & 83 & \textbf{Social} (Status, Gender, Authority, Interaction) \\
\hline
2429 & 79 & \textbf{mRNA} (Transcript, mRNA, Signal, Frequency, Expression) \\
\hline
1868 & 77 & \textbf{Literature} (Citation, Publication, Journal, Literature) \\
\hline
1349 & 67 & \textbf{Assay} (Method, Assay, Type, Measurement) \\
\hline
6241 & 62 & \textbf{Genetic} (Strain, Allele, Mutation, Genotype, SNP) \\
\hline
6052* & 50 & \textbf{Enzyme} (EC, Enzyme, Reaction) \\
\hline
3524* & 50 & \textbf{Phenotype} (Phenotype, Omim) \\
\hline
2490 & 50 & \textbf{Organization} (Component, Complex, Assembly, Entity)\\
\hline
1849 & 49 & \textbf{Interaction} (Interaction, Experiment, Strength)\\
\hline
2897 & 49 & \textbf{Cell Line} (Cell, Line, Tissue, Type, Node)\\
\hline
6283 & 39 & \textbf{Organism} (Organism, Host, Entity)\\
\hline
2911 & 38 & \textbf{Ontology} (Synonym, Term, Ontology) \\
\hline
1787 & 32 & \textbf{Evolution} (Member, Family, Orthologue, Homologue, Evidence) \\
\hline
5695 & 31 & \textbf{Literature} (Pubmed, Keyword)\\
\hline
\hline
\end{tabular}
\caption[Primary composition of communities in terms of key biomedical concepts]{Primary composition of communities in terms of key biomedical concepts. (*) marked communities are visualized in \textbf{Figure \ref{fig:labelmismatchcomm}}.}
\label{tab:communitycompo}
\end{table*}

Smaller communities ($<10$ members) also consist of specific schema elements in different sources that are similar to each other but that are in label mismatch through use of different URIs (e.g., \textsc{snoRNA} is represented using \\ \texttt{MOLD:yeastmine\_SnoRNAGene}, \texttt{MOLD:humanmine\_SnoRNA}, \texttt{EBIRDF:ensembl/snoRNA}, \\ \texttt{BIO2RDF:ncbigene\_vocabulary/SnoRNA-Gene}, \texttt{NBDC:mbgd\#snoRNA}). Knowledge of such communities will aid biomedical researchers to determine which sources in the LSLOD cloud contain information relevant to their domain of interest (e.g., research on small nucleolar RNAs \textsc{snoRNA}, a class of small RNA molecules that primarily guide chemical modifications). 

\section*{Discussion}
\label{sec:discussion}
Semantic heterogeneity across different data and knowledge sources, either in the use of varying schemas or entity notations, severely hinders the retrieval and integration of relevant data and knowledge for use in downstream biomedical applications (e.g., drug discovery, pharmacovigilance, question answering). Enhancing the quality and decreasing the heterogeneity of biomedical data and knowledge sources will lead to enhanced inter-disciplinary biomedical research, which will lead to developing better clinical outcomes and increasing our knowledge on biological mechanisms. 

So far, significant resources have been invested to represent, publish, and link biomedical data and knowledge on the Web using Semantic Web technologies to create the Life Sciences Linked Open Data (LSLOD) cloud. However, there are very few biomedical applications that query and use \emph{multiple} linked data sources and biomedical ontologies directly on the Web to generate new insights, or to discover novel implicit associations \emph{serendipitously}. This is, in part, due to the semantic heterogeneity emanating from underlying sources, which has also penetrated the LSLOD cloud in varying forms. Some of the primary benefits of the LSLOD cloud --- querying, retrieval, and integration of heterogeneous data and knowledge from multiple sources on the Web seamlessly --- cannot be availed. 

Linked data publishers are strongly encouraged to reuse existing models and vocabularies and to provide links to other sources in the LSLOD cloud \cite{bizer2009linked,marshall2012emerging}. Similarly, biomedical ontology developers are strongly encouraged to reuse equivalent classes existing in other ontologies, while building new ontologies \cite{simperl2009reusing,corcho2003methodologies}. There are several benefits if biomedical ontologies and linked data sources reuse content from existing sources --- \textit{i)} reduction in data and knowledge engineering and publishing costs, \textit{ii)} decrease in semantic heterogeneity and increase in semantic interoperability across datasets, and \textit{iii)} ease of querying, retrieval, and integration of data from multiple sources simultaneously through existing query federation methods \cite{kamdar2017systematic}. On the other hand, lack of reuse will manifest as increased semantic heterogeneity across different LSLOD sources.

It can be asserted through the findings presented in this research that the Life Sciences ``Linked'' Open Data cloud is not actually as densely connected as is often indicated through the ubiquitously-referenced LSLOD cloud diagram. Several sources, which may exhibit a higher degree of intra-linking are not inter-linked with other RDF graphs and exist as stand-alone data sources converted using RDF (\textbf{Figure \ref{fig:interlinkLD}}). Similarly, \textbf{Figure \ref{fig:VocabularyReuse}} shows that several biomedical linked data sources use unpublished schemas. Biomedical linked data sources do reuse schema elements from a set of popular vocabularies. However, these vocabularies originate from the Semantic Web community (e.g., those that are mainly used to refer to data elements and attributes) rather than from the biomedical community (e.g., those that are used to provide normalization schemes for biomedical entities). Hence, the elements of these vocabularies are not useful for data integration and integrated querying from a biomedical perspective (e.g., for retrieving all \textsc{Drug--Protein Target Interactions}). 

\subsection*{Similar Content across the LSLOD Cloud}
As seen in \textbf{Figure \ref{fig:labelmismatchcomm}}, several online sources have data and knowledge pertaining to \textsc{Enzyme} or \textsc{Phenotype} entities but use completely different schema elements. Communities of similar content depicted in \textbf{Figure \ref{fig:labelmismatchcomm}}, while showcasing the different schema element URIs to represent similar content, may also provide an idea on the different equivalence attribute URIs used to link entities across different sources to a specific entity coding scheme. For example, the \textsc{Enzyme} entities in the different sources of the LSLOD cloud are linked to the Enzyme Commission number (a numerical classification scheme for enzymes) using different attribute URIs with labels such as \textsc{Ec Code}, \textsc{Ec Number}, \textsc{X Ec}, and \textsc{Has Ec Number}. Similarly, \textsc{Phenotype} entities are linked to the OMIM phenotype identifiers (i.e., Online Mendelian Inheritance in Man is a knowledge base of human genes and genetic phenotypes \cite{hamosh2005online}) using different attribute URIs with labels such as \textsc{Omim Reference}, \textsc{X Omim}, and \textsc{Has Omim Phenotype}. Communities that are composed entirely of different attribute URIs referring entities in different sources to a uniform coding scheme are also observed (e.g., \textsc{Ensembl} and \textsc{HGNC} Gene Coding scheme attributes). The knowledge on the different equivalence attribute URIs could guide query federation methods to perform entity reconciliation after the retrieval of information from multiple sources. 

Through a manual inspection of such ``identifier'' communities (e.g., communities 1417 and 1632 in \textbf{Table \ref{tab:communitycompo}}), we were able to detect different URI representations for similar entities (classes and instances). A few examples of different URI representations are shown in \textbf{Table \ref{tab:lduriintent}} for chemical entities of ChEBI \cite{hastings2013chebi}, PubChem \cite{fu2015pubchemrdf} and MeSH \cite{bushman2015transforming} sources. Most linked data querying architectures rely on the existence of exact URIs to query multiple sources simultaneously (using data warehousing, link traversal or query federation methods) \cite{saleem2014big,kamdar2019enabling,kamdar2017phlegra}. However, since these URIs are essentially different, querying architectures will not be able to integrate data and knowledge from different sources (e.g., protein--ligand binding information from PDB and biological assay experimental data from PubChem). These examples qualify as \textbf{``intent for reuse''} and add to the challenges of semantic heterogeneity across the LSLOD cloud.

The use of word embeddings in the community detection algorithm enables the identification of mappings between similar linked data schema elements (e.g., \texttt{kegg:therapeutic-category} $\leftrightarrow$ \texttt{drugbank:Drug-Classification-Category}: 0.93, \texttt{obo:regulates} $\leftrightarrow$ \texttt{drugbank:mechanism-of-action}: 0.75, \texttt{faldo:location} $\leftrightarrow$ \texttt{drugbank:cellular-location}: 0.89). These mappings may provide suggestions to linked data publishers and consumers on relevant information in different sources. However, many of these mappings may not be accurate due to reliance on the word embeddings, as well as the 0.75 threshold on the cosine similarity scores. For example, as seen in \textbf{Figure \ref{fig:labelmismatchcomm}}, while the \textsc{Reaction} concept in the \textsc{Enzyme} community symbolizes biochemical reactions where enzyme entities may act as catalyzing agents, an error with the community detection algorithm also indicates that \textsc{Adverse Reaction} (clinical concept observed in patients treated with a set of drugs) may be a part of this community. This is due to the fact that the singular schema element pertaining to \textsc{Adverse Reaction} from an RDF graph in the Linked Life Data project \cite{linkedlifedatalink} is linked to schema elements pertaining to \textsc{Biochemical Reaction} (i.e., the edges have a cosine similarity score of more than 0.75).  

\subsection*{Intent for Reuse across the LSLOD Cloud}
\label{sec:intent}
\begin{table*}[!htb]
\centering
\footnotesize
\renewcommand{\arraystretch}{1.5}
\begin{tabular}{l|l|l}
\hline
\textbf{Entity Origin} & \textbf{URI Representation} & \textbf{Sources}  \\
\hline
\hline
\textbf{ChEBI} & \texttt{http://purl.obolibrary.org/obo/CHEBI/}* & LinkedSPL, PubChem, NBDC \\
& & ExpresssionAtlas, BioSamples \\
& \texttt{http://identifiers.org/chebi/CHEBI:} & BioModels, PDB, WikiPathways \\ & & Bio2RDF, PathwayCommons \\
& \texttt{http://identifiers.org/obo.chebi/CHEBI:} & BioModels, BioPax, WikiPathways \\
& \texttt{http://www.ebi.ac.uk/chebi/} & PDB, ChEMBL\\
& \texttt{http://bio2rdf.org/chebi:} & Bio2RDF  \\ 
\hline
\textbf{PubChem} & \texttt{http://identifiers.org/pubchem.compound/}  & Bio2RDF, WikiPathways, NBDC \\
\textbf{Compound} & \texttt{http://rdf.ncbi.nlm.nih.gov/pubchem/compound/}* & WikiPathways, PubChem \\
& \texttt{http://bio2rdf.org/pubchem.compound:} & Bio2RDF \\
& \texttt{http://pubchem.ncbi.nlm.nih.gov/compound/}  & NBDC \\
\hline
\textbf{MeSH} & \texttt{http://purl.bioontology.org/ontology/MESH/}* & NBDC, PDB, LinkedSPL \\
& \texttt{http://id.nlm.nih.gov/mesh/{YEAR}/}* &  MeSH, DisGenet, PDB \\
& \texttt{http://identifiers.org/mesh/} &  Bio2RDF, DisGenet, PDB\\
& \texttt{http://www.ncbi.nlm.nih.gov/mesh/} & BioSamples \\
& \texttt{http://rdf.imim.es/rh-mesh.owl\#} &  DisGenet\\
& \texttt{http://bio2rdf.org/mesh\_resource:} &  Bio2RDF \\
& \texttt{http://bio2rdf.org/mesh:} &  Bio2RDF \\
\hline
\hline
\end{tabular}
\caption[Different kinds of URI representations observed in LD Sources]{Different kinds of URI representations for chemical compounds observed in LD Sources. (*) marks the recommended representation(s).}
\label{tab:lduriintent}
\end{table*}
A few linked data sources do reuse content from existing biomedical ontologies, but in many cases the reuse is limited to very few classes or involve incorrect representations. Previously, we have documented similar cases to be \textbf{``intent for reuse''} on the part of ontology engineers in the biomedical domains \cite{kamdar2017systematic}. \textbf{Table \ref{tab:lduriintent}} documents similar cases of \textbf{intent for reuse} that were observed across the biomedical linked data sources. Essentially, these URI patterns generally have the same identifier and source ontology or vocabulary, but are reused from different versions of the source ontology or vocabulary, or represented using different notations or namespaces. These patterns cannot be considered as actual reuse, as these are different, and often incorrect, representations for the same terms, and no explicit mappings are found. Hence, the advantages of reuse cannot be experienced. By using the correct representations, the semantic heterogeneity (i.e., content overlap) can be reduced substantially.  However, ontology engineers and linked data publishers may lack sufficient guidelines and tools to perform these tasks. These examples severely hinder most query federation and link traversal methods for heterogeneous data and knowledge integration on the Life Sciences Linked Open Data cloud. 

For a more concrete example on how intent for reuse and semantic heterogeneity hampers link traversal and query federation across the LSLOD cloud, consider the user query \textit{``Retrieve Assays which identify potential Chemopreventive Agents that have Molecular Weight < 300 g/mol and target Estrogen Receptor present in Human.''} This query requires the retrieval of knowledge from the Kyoto Encyclopedia of Genes and Genomes (KEGG) knowledge base \cite{kanehisa2000kegg} on all the chemicals that target \textsc{Estrogen Receptor} protein, and then the retrieval of assay data on these chemicals from the ChEMBL chemical bioactivity repository \cite{willighagen2013chembl}. The KEGG knowledge base and the ChEMBL data repository are both published on the LSLOD cloud \cite{callahan2013bio2rdf,jupp2014ebi}, and have chemical URIs that have \texttt{x-ref} links to chemical URIs in the ChEBI ontology of biological chemicals \cite{hastings2013chebi}. However, since Bio2RDF KEGG uses the URI representation scheme \texttt{http://bio2rdf.org/chebi:*} and ChEMBL uses the URI representation scheme \texttt{http://purl.obolibrary.org/obo/CHEBI/*} for the same identifiers in the ChEBI ontology (\textbf{Table \ref{tab:lduriintent}}), it is impossible to formulate a federated SPARQL query that can query KEGG and ChEMBL, simultaneously and retrieve the relevant results in an integrated fashion.

Increasing the reuse of schema elements from existing vocabularies and ontologies may decrease semantic heterogeneity and enable the integrated querying of multiple data and knowledge sources across the LSLOD cloud using the Semantic Web technologies. In cases where linked data publishers have the need to use a custom vocabulary or ontology, the overarching advice would be to publish this vocabulary or ontology to a popular repository, so other publishers can use these sources. 

\section*{Conclusion}
Semantic Web and linked data technologies garner significant interest and investment from the biomedical research community toward tackling the diverse challenges of heterogeneous biomedical data and knowledge integration. In this paper, we conduct an empirical meta-analysis to demonstrate that there is considerable heterogeneity and discrepancy in the quality of Life Sciences Linked Open Data (LSLOD) sources on the Web, which are published using the above technologies. We discuss on whether the LSLOD cloud can truly be considered \textit{``linked''} due to the heterogeneous schemas, varying entity notations, the lack of mappings between similar entities, and the lack of reuse of common vocabularies. While increasing reuse across biomedical ontologies and linked data sources is definitely considered to be an attractive alternative to decrease semantic heterogeneity, it will take a while for biomedical ontology developers and linked data publishers to realize the incentives of reuse and actually embrace the guidelines and best practices while publishing data and knowledge. Actual reuse through the use of correct URI representations in biomedical ontologies and linked data sources, rather than \textit{intent to reuse}, can decrease semantic heterogeneity substantially itself. The findings from our meta-analysis across the LSLOD cloud, as well as the resources such as the LSLOD schema graph made available through this research, will lead to the development of better tools and methods to increase vocabulary reuse and enable efficient querying and integration of multiple heterogeneous biomedical sources on the Web.

\bibliography{references}


\section*{Acknowledgements}
The authors would like to dedicate this paper to the memory of Dr. Amrapali Zaveri, a staunch advocate for data quality issues and the development of crowd-sourcing and human-in-the-loop methods to improve the quality of several data sources on the Web, who passed away in January 2020. The authors would also like to acknowledge Michel Dumontier, Javier Fern{\'a}ndez, Rafael Gonçalves, Matthew Horridge, Marcos Martinez, Csongar Nyulas, Axel Polleres, and Tania Tudorache for valuable feedback on this research. M.K. and M.M. were supported in part by grants GM086587, GM103316, and HG004028 from the US National Institutes of Health.

\section*{Author contributions statement}
M.K. conducted the meta-analysis of data sources across the Life Sciences Linked Open Data cloud.  All authors reviewed and approved the manuscript. 

\section*{Competing Interests}
The Authors declare no Competing Financial or Non-Financial Interests.

\section*{Data Availability}
The LSLOD Schema Graph extracted from the different SPARQL endpoints of the biomedical linked data graphs in the LSLOD cloud is made available in the JavaScript Object Notation (JSON) format serialized as a Pickle file in the FigShare repository (DOI: \url{https://doi.org/10.6084/m9.figshare.12402425.v2}) under a CC BY 4.0 license \cite{kamdar_2020}. The project also contains tab-separated values files consisting of information about extracted classes, object properties, data properties, datatypes, as well as of statistics on the linked data graphs from the LSLOD cloud used in this research.

\section*{Code Availability}
The different visualizations and observations presented in this research are also made available at the project website (\url{http://onto-apps.stanford.edu/lslodminer}). The scripts for the extraction algorithm and the community detection algorithm used in this research are made available on GitHub (\url{https://github.com/maulikkamdar/LSLODQuery}).

\section*{Supplementary Material}
The Extraction algorithm uses a set of SPARQL query templates (\textbf{Listing \ref{lst:sparqlqset}}) to extract schemas (e.g., classes, properties, domains, ranges) as well as sample instances of the classes and their property values from the SPARQL endpoints in the LSLOD Cloud. The input of the Extraction algorithm is the SPARQL endpoint and its version, which can be automatically detected from a SPARQL endpoint based on the queries and keywords supported. These schema elements are further annotated with few characteristics that their instances can undertake (e.g., namespaces, regular expression patterns of the instance URIs, type of attribute datatypes). The Extraction algorithm is presented in \textbf{Algorithm 1}.

\begin{lstlisting}[captionpos=b, caption={SPARQL Query templates to extract schema descriptions from the LSLOD Cloud}, label=lst:sparqlqset, basicstyle=\linespread{1}\ttfamily,frame=single]
------------------ SPARQL QUERY 1 ------------------
SELECT DISTINCT ?g WHERE {
	GRAPH ?g { ?s ?p ?o }
}

------------------ SPARQL QUERY 2 ------------------
SELECT ?Concept (COUNT (?x) AS ?cCount) WHERE {
	GRAPH <GRAPH_URI> { ?x rdf:type ?Concept }
} GROUP BY ?Concept ORDER BY DESC(?cCount)

------------------ SPARQL QUERY 3 ------------------
SELECT DISTINCT ?p ?c (COUNT(?x) AS ?count) ?valType WHERE {
	GRAPH <GRAPH_URI> { ?x rdf:type <CONCEPT_URI>; ?p ?o . 
    OPTIONAL {?o rdf:type ?c} . 
    FILTER(!(?p = 'rdf:type')) . 
    BIND(DATATYPE(?o) AS ?valType) }
} GROUP BY ?p ?c ?valType ORDER BY DESC(?count)

------------------ SPARQL QUERY 4 ------------------
SELECT ?x WHERE {
	GRAPH <GRAPH_URI> { ?x rdf:type <CONCEPT_URI> }
} ORDER BY RAND() LIMIT 2000

------------------ SPARQL QUERY 5 ------------------
SELECT ?x WHERE {
	GRAPH <GRAPH_URI> { ?c rdf:type <CONCEPT_URI>; <PROPERTY_URI> ?x }
} ORDER BY RAND() LIMIT 2000
\end{lstlisting}

As shown in \textbf{Algorithm 1}, the \textbf{extractor} decides on the set of SPARQL query templates to execute against the input SPARQL endpoint given its version (\textbf{Step 2}). The extractor queries the SPARQL endpoint and extracts the set of graph URIs for all named RDF Graphs $\symNG$ exposed through the SPARQL endpoint (\textbf{Step 3, SPARQL Query $\mathbf{SQ_1}$}). For each graph URI, the extractor extracts the set of classes $\symC$ and the total number of instances for each class $k(\symC)$ (\textbf{Step 7, $\mathbf{SQ_2}$}). The extractor identifies the properties $\symP$ for which a specific class serves as the domain for all assertions, and also retrieves the range classes $r(\symP)$ for object properties, or the datatypes $dt(\symP)$ for data properties (\textbf{Step 12, $\mathbf{SQ_3}$}). The Extraction algorithm also extracts a random sample (without replacement) of 2,000 instances for each class (\textbf{Step 9, $\mathbf{SQ_4}$}) and 2,000 property assertions (\textbf{Step 14, $\mathbf{SQ_5}$}) for a class--property combination, and generates summary characteristics (e.g.is\_categorical, datatype, namespace, median length) of the data.

The Extraction algorithm accounts for different versions of SPARQL supported at the remote SPARQL endpoints using a `version' parameter input $v(\symSE)$. Depending on the SPARQL version, certain SPARQL constructs such as \textsc{Bind} and \textsc{Group By} may not be supported, and alternative SPARQL queries are formulated (not shown here). Alternative SPARQL queries are also formulated if the remote endpoint times out while query processing. 

Classes, properties, and datatypes, are represented as nodes in the schema graph $\symG_{{\symT}-Box}$ extracted by the Pattern Miner (\textbf{Steps 11, 17, 18, 21, 22}). The edges link property nodes to class nodes and datatype nodes depending on the domain and range descriptions of the corresponding property (\textbf{Steps 19 and 23}). The nodes and edges are annotated by the count ($k(\symC)$  and $k(\symP)$) and summary characteristics of corresponding instances or property assertions respectively (\textbf{Steps 11, 19, 23}). This schema graph is a fragment of the actual $\symT$-Box of the RDF graph. The $\symG_{{\symT}-Box}$ graphs extracted for all RDF named graphs $\symNG$ in the LSLOD cloud are merged to generate the schema graph $\symG_{LSLOD}$ of the entire LSLOD cloud.


\begin{algorithm}
\setstretch{1}
	\label{ref:algo1}
    \SetKwInOut{Input}{Input}
    \SetKwInOut{Output}{Output}
    \underline{function \textsc{Extractor}} $(\symSE, v(\symSE))$\;
    \Input{A SPARQL Endpoint URI $\symSE$ and the supported version $v(\symSE)$}
    \Output{A dictionary of (graph URIs, schema ($\symT$-Box) fragments, each represented as a graph $\symG_{{\symT}-Box}$), extracted from the SPARQL endpoint $\symSE$}
    \tcc{Generate SPARQL Queries given the version of the endpoint (Listing \ref{lst:sparqlqset})}
    $\{SQ_i|i=1,2,\ldots,5\} \leftarrow $\textsc{SPARQL\_Queries}($v(\symSE)$)\;
    \tcc{Extract all the Graph URIs}
    $\{URI_g\} \leftarrow SQ_1$ $(\symS\symE)$  \;
    \tcc{An empty dictionary to store the Graph URIs and T-Box schema fragments}
    $SF \leftarrow \{\}$ \;
    \For{$uri_g \in \{URI_g\}$}{
    $\symG_{{\symT}-Box} \leftarrow \varnothing $ \;
    $\symC, k(\symC) \leftarrow SQ_2(\symS\symE, uri_g)$ \;
    \For{$c \in \symC$} {
    \tcc{Extract a randomized sample of 2000 instances}
    $\symA_{(c)-2000} \leftarrow SQ_4(\symSE, uri_g, c)$\;
    \tcc{Generate a data summary of the instances}
    $sum_c \leftarrow $\textsc{Summarize}($\symA_{(c)-2000}$) \;
    Add node to $\symG_{{\symT}-Box}$ with URI $c$, Type $class$, Count $k(c)$ and Summary $sum_c$ \;
    $\symP, r(\symP), dt(\symP), k(\symP) \leftarrow SQ_3(\symS\symE, uri_g, c)$ \;
    \For{$p \in \symP$} {
    \tcc{Extract a randomized sample of 2000 assertions}
    $\symA_{(c,p)-2000} \leftarrow SQ_5$  $(\symSE, uri_g, c, p)$\;
    \tcc{Generate a data summary of the assertions}
    $sum_{cp} \leftarrow $\textsc{Summarize}($\symA_{(c)-2000}$) \;
    \tcc{Data property if range URI is missing, else Object property}
    \eIf{$r(p) = \varnothing$}
    {
    	Add node to $\symG_{{\symT}-Box}$ with URI $p$, Type $data\;property$\;
        Add node to $\symG_{{\symT}-Box}$ with URI $dt(p)$, Type $datatype$\;
    	Add edges $c\rightarrow p$ and $p \rightarrow dt(p)$ to $\symG_{{\symT}-Box}$, with Count $k(p)$ and Summary $sum_{cp}$\;
	}
    {
    	Add node to $\symG_{{\symT}-Box}$ with URI $p$, Type $object\;property$\;
        Add node to $\symG_{{\symT}-Box}$ with URI $r(p)$, Type $class$\;
    	Add edges $c\rightarrow p$ and $p \rightarrow r(p)$ to $\symG_{{\symT}-Box}$, with Count $k(p)$ and Summary $sum_{cp}$\;
    }
    
    }
    }
    $SF[uri_g] \leftarrow \symG_{\symT-Box}$ \;
    }
    return $SF$\;
    \caption{LSLOD Schema Extraction Algorithm}
\end{algorithm}

\end{document}